\documentclass[10pt,twocolumn,letterpaper]{article}

\usepackage[pagenumbers]{cvpr} % To force page numbers, e.g. for an arXiv version
\usepackage[ruled]{algorithm2e}
\usepackage{tikz}

\definecolor{gold}{rgb}{0,0,1.0}
\definecolor{silver}{rgb}{0.0666,0.3333,0.8}
\definecolor{bronze}{rgb}{0.6235,0.7725,0.910}
\usepackage{siunitx}
\usepackage{graphicx}
\definecolor{cvprblue}{rgb}{0.21,0.49,0.74}
\usepackage[pagebackref,breaklinks,colorlinks,citecolor=cvprblue]{hyperref}
\usepackage{pgfplots}
\usepackage{pgfplotstable}

\def\paperID{6911} % *** Enter the Paper ID here
\def\confName{CVPR}
\def\confYear{2024}

\title{TSDF-Sampling: Efficient Sampling for Neural Surface Field using Truncated Signed Distance Field}

\author{Chaerin Min$^{1,2*\dag}$ \quad
Sehyun Cha$^{1,3*\dag}$ \quad
Changhee Won$^1$ \quad
Jongwoo Lim$^{1,4\dag}$\\
$^1$MultiplEYE \quad
$^2$Brown University \quad
$^3$Hanyang University \quad
$^4$Seoul National University\\
\small{$^*$ denotes equal contribution}
}

\begin{document}
\twocolumn[{
\maketitle
\setlength{\parindent}{3pt}
\begin{center}
    \renewcommand{\tabcolsep}{1.5pt}
    \centering \small
    \begin{tabular}{cccccccc}
    % &&\multicolumn{5}{*}{$\rightarrow$}\\
    &&
    % \normalsize Input& 
    \multicolumn{5}{c}{Average number of samples per ray}\\
    &&96&28&24&19&13  \vspace{-.5em}\\
    &&
    \multicolumn{5}{c}{
    \begin{tikzpicture}
        \draw[->, line width=0.8mm] (0,0) -- (16.5,0);
    \end{tikzpicture}} \vspace{-.0em}\\
    \rotatebox{90}{\,\, Hierarchical}&\rotatebox{90}{\quad sampling}&
    \includegraphics[width=0.185\textwidth]{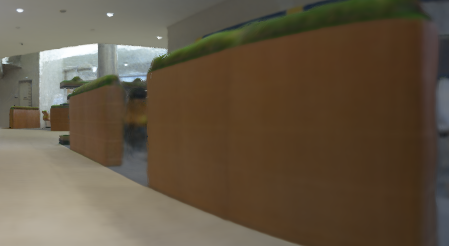} &
    \includegraphics[width=0.185\textwidth]{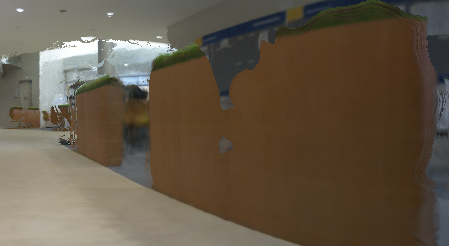} &
    \includegraphics[width=0.185\textwidth]{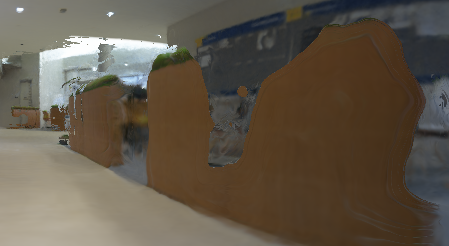} &
    \includegraphics[width=0.185\textwidth]{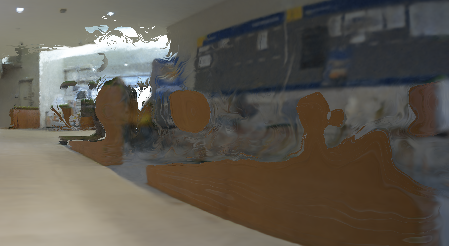} &
    \includegraphics[width=0.185\textwidth]{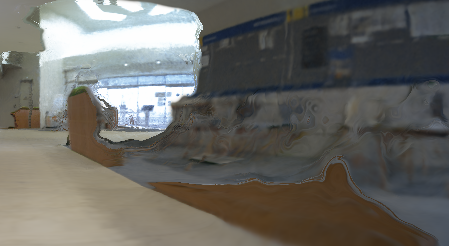} \\
    % Input image &
    % \multicolumn{5}{c}{(a) MonoSDF\cite{monosdf}}\\
    \rotatebox{90}{$\quad\;$Ours}&&
    \includegraphics[width=0.185\textwidth]{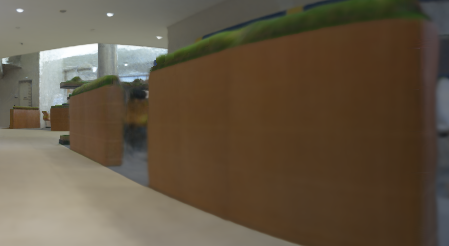} &
    \includegraphics[width=0.185\textwidth]{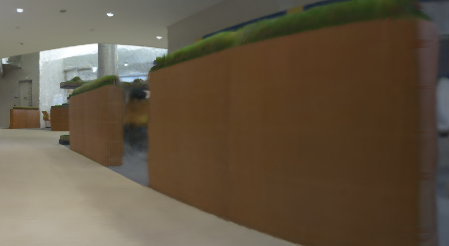} &
    \includegraphics[width=0.185\textwidth]{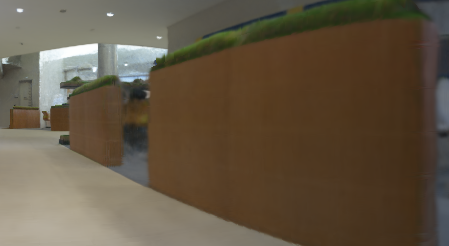} &
    \includegraphics[width=0.185\textwidth]{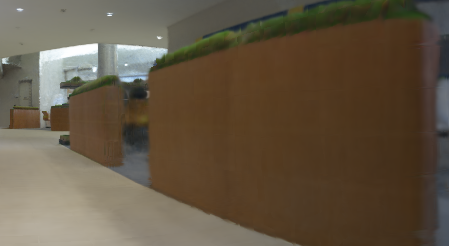} &
    \includegraphics[width=0.185\textwidth]{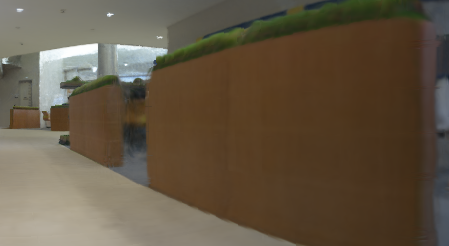} \vspace{-.5em}\\
    % Input image&
    % \multicolumn{5}{c}{(b) Ours}
    \end{tabular}
    \captionof{figure}{The novel view inference of our method is shown (cropped from 360-degree panorama images). 
    As the number of samples decreases, our rendering quality remains intact, while the hierarchical sampling used in \cite{nerf,neus,monosdf} fails to detect relatively thin structures, such as the partitions. 
    % Note that, to further challenge our generalization abilities, we captured the train dataset with UW fisheye cameras and tested it with a completely different camera model, i.e., Equirectangle, in this figure. 
    % The images has been cropped around the center to better visualize the details.
    }
    \label{fig:teaser}
\end{center}

}]

\footnotetext[1]{$\dag$ This work was done when they were at Hanyang University.}
\begin{abstract}

% abstract version 1

% We see an explosion of the Neural Radiance Fields (NeRFs) literature, for their compelling result of novel view synthesis. Recently, Neural Implicit Surfaces achieved impressive dense surface reconstruction by leveraging the continuity of MLPs to approximate SDFs. Such Neural Implicit Surfaces, however, tend to require larger networks and exhaustive samplings to obtain both photo-realistic images and high-fidelity depth map results, ended up with slower rendering. On the other hand, recent fast NeRF approaches, which are mainly based on density fields, leave geometric accuracy an unexplored area. In this paper, we present TSDF-NeRF, which is, to the best of our knowledge, the first approach to carry the performance of the state-of-the-art Neural Implicit Surfaces, while substantially boosting their speed. In effect, our TSDF-NeRF can provide acceleration to any models that employ the volume rendering scheme. Our method incorporates traditional TSDF integration to set reliable bounds of the sampling complexity, without requiring any additional sensor depths. In addition, with our different sampling bounds for each ray, our method adaptively determine the number of sufficient samples of each ray. We demonstrate that this approach performs on par with the state-of-the-art baselines, with up to 1300\% faster rendering time. Furthermore, together with our recovery algorithm, we show that our model robustly accelerates the pipeline in large real-world scenes. Our code and project page will soon be available at \href{https://www.naver.com/}{https://github.io/tsdf-nerf}.

\vspace{-1em}
% Abstract version 2. 
Multi-view neural surface reconstruction has exhibited impressive results. However, a notable limitation is the prohibitively slow inference time when compared to traditional techniques, primarily attributed to the dense sampling, required to maintain the rendering quality. This paper introduces a novel approach that substantially reduces the number of samplings by incorporating the Truncated Signed Distance Field (TSDF) of the scene. 
While prior works have proposed importance sampling, their dependence on initial uniform samples over the entire space makes them unable to avoid performance degradation when trying to use less number of samples. 
% While prior works have sought acceleration by storing  density in grid-based data structures, the inherent discrete nature of these grids often leads to over-fitting and artifacts. 
% In contrast, our method remains to be a continuous representation, thereby ensuring high quality.
In contrast, our method leverages the TSDF volume generated only by the trained views, and it proves to provide a reasonable bound on the sampling from upcoming novel views. 
As a result, we achieve high rendering quality by fully exploiting the continuous neural SDF estimation within the bounds given by the TSDF volume.
% As a result, within the bound, finding the accurate surface becomes an easier problem.
% In addition, with the assistant of the TSDF volume, we are able to carry the advantage of neural approximation of SDF as a continuous representation. 
% % keep the rendering from the neural surface field continuous. 
% , even with an extremely small number of sample points.
Notably, our method is the first approach that can be robustly plug-and-play into a diverse array of neural surface field models, as long as they use the volume rendering technique.
Our empirical results show an 11-fold increase in inference speed without compromising performance.  The result videos are available at our project page: \href{https://tsdf-sampling.github.io/}{https://tsdf-sampling.github.io/}
% our method is able to perform on par with the state-of-the-art, only with 13\% number of samplings. 
% Our code and project page will be available at
% \href{https://www.naver.com/}{https://tsdf-sampling.github.io/}.

\end{abstract}   
\section{Introduction}
\label{sec:intro}

\indent 3D reconstruction from multi-view images \cite{multiview1,multiview2,early3,poxels} is a key challenge in computer vision, robotics, and graphics. Early works \cite{early1, early3, poxels,point3} relied on explicit representations including point clouds \cite{point1,point2,point3,point4} and voxels \cite{marchingcubes,voxel1,voxel2,voxel3,voxel4}. 
However, point clouds and voxels are limited to being sparse and discrete, respectively. 
% Recently, Neural surface fields with differentiable rendering \cite{nerf} mitigated these problems to some extent, improving the performance. However, neural surface fields are painfully slow. VolSDF \cite{volsdf},  NeuS \cite{neus}, and their variants \cite{neuralangelo,neus2,geoneus} are those of the most popular neural surface approaches, but their inference time for a FHD image are more than 10 seconds on a GPU. 
Neural surface fields with differentiable rendering, such as VolSDF \cite{volsdf}, NeuS \cite{neus}, and their variants \cite{neuralangelo, neus2, geoneus} have alleviated these issues, yet they suffer from painfully slow inference, with more than ten seconds for a FHD image on a GPU. 

Most of the neural surface reconstruction frameworks \cite{unisurf, deepsdf,monosdf,volsdf,neus,nerf,mipnerf,ngp,nglod} cast rays, on which points are sampled and then passed to the MLPs \cite{mlp}. Since MLPs consume most of the reconstruction time, the overall time complexity of neural surface fields can be summarized as follows: $O(mN)$, where $N$ is the image resolution (a given parameter), and $m$ is the average number of samples per ray, often manually specified~\cite{nglod,volsdf}. Therefore, our research question is as follows: Is it possible to reduce $m$ without degrading the rendering quality of the 3D neural SDFs? 

To our surprise, with the conventional sampling strategy, the answer was \textit{No}, shown in Sec. \ref{sec:exp}. 
% With the conventional sampling strategy, the answer is \textit{No}, according to our investigation.
% Then, we propose a novel approach that can decrease the m without impairing the rendered results of 3D reconstruction. 
First, we successfully avoid the speed-quality tradeoff by eliminating most unnecessary sampling.
% We further ensure the safety of such elimination with our recovery technique.    
% previous methods typically trade off between the speed and quality. 
The existing methods \cite{mipnerf,neus,manhattansdf,nerfinthewild,neuralangelo} use the importance sampling of \cite{nerf} and rely heavily on the initial (coarse) sampling, because they do not know the space occupancy without querying MLPs. 
Therefore insufficient initial sampling leads directly to poor rendering quality.
% Therefore, limiting the exhaustive initial sampling and not failing significantly is unrealistic. 
% and is capable of reducing the m significantly without quality decrease. 

We break through this limitation by incorporating the traditional Truncated Signed Distance Field (TSDF) \cite{tsdf,kinectfusion} integration technique with the volume rendering \cite{nerf}. 
Essentially, our method adaptively sets reasonable range bounds of sampling per individual ray, around which the surface is likely to be located. 
The number of samples is determined by the distribution of the SDF values along each ray on the TSDF grid created at the training time.
We also double-check and handle the incorrect bounds by TSDF for flawless rendering. 
% Furthermore, the chance of having a surface within the bound is double ensured by our recovery algorithm. 
As a result, our adaptive sampling allows for more sampling in difficult rays and less sampling in easier rays, thus minimizing unnecessary sampling. 
% In our algorithm the number of samples is dependent on each ray's sampling range, which is determined by the distribution of the SDF values along each ray on the TSDF grid, that is created solely by the trained model.

% Second, each prior method \cite{monosdf,volsdf,neus,neus2,neuralangelo,deepsdf,manhattansdf,unisurf,neuralrecon3dinthewild,raydf} proposed framework that is difficult to be mounted on other previous or future works. 
Second, our proposed method is model-agnostic, and it can be readily applied to improve the speed of any other neural surface fields methods \cite{monosdf,volsdf,neus,neus2,neuralangelo,deepsdf,manhattansdf,unisurf,neuralrecon3dinthewild,raydf} without having to train them again, as long as they use volume rendering \cite{volumerender} for their inference of the 3D representation. 
Our exhaustive analysis on the reason for the efficacy of our method is described in Sec. \ref{sec:method} and \ref{sec:exp}. 
In the $\bold{Lobby}$ dataset of 20$\times$20$\times$5 $\text{m}^3$, we achieve accurate inference performance with as little as 12 samples per ray in average. 
Finally, our technical contributions are summarized as follows: 
\begin{itemize}
    \item 
    We introduce the TSDF Sampling, the first method that significantly accelerates the rendering of continuous neural SDFs with complex geometry in a reasonable inference time.  
    \item 
    By using the classical TSDF \cite{tsdf} technique and exploiting the inherent geometry encoded in the neural fields, our method is able to find the appropriate range bound per ray and query only the necessary samples in the bound. 
    Rare incorrectly estimated bounds are explicitly handled to ensure the rendering quality.
    % By using the classical TSDF \cite{tsdf} technique and exploiting the inherent geometry that has been implicitly encoded in the neural fields, our method is capable of querying only the samples that are effective enough to the projected result. 
    % we improve upon the existing methods of sampling the continuous volume rendering, when only given the trained model.
    % \item  We propose our further adaptive method and recovery algorithm, which empowers more robust generalization to large real-world scenes.
    \item 
    Our algorithm is model-agnostic and can be easily plugged into different neural surface field models. 
    Extensive experiments on different datasets and networks prove this.
    
    % easily generalizes across different 00
\end{itemize}

\section{Related Works}
\label{sec:relatedwork}

% We shhould explicitly mention other adaptive sampling approaches. Include and describe as many works as possible. 
% \subsection{Neural Fields}
% Novel view synthesis is a task of render images from unboserved viewpoints. Classical methods obtain the view synthesis from interpolation of light fields. Lately, NeRF[] and its variants[] showed notable results on photo-realistic renderings. Follow-up works extended this concepts into various applications; [] used canonicalization that effectively models dynamic object and [] relaxed the photometric contraint that often fails in different lighting effects. 
\begin{figure*}[!tp]
    \renewcommand{\tabcolsep}{5.0pt}
    \centering \small
    \begin{tabular}{cc}
        \includegraphics[width=0.43\textwidth]{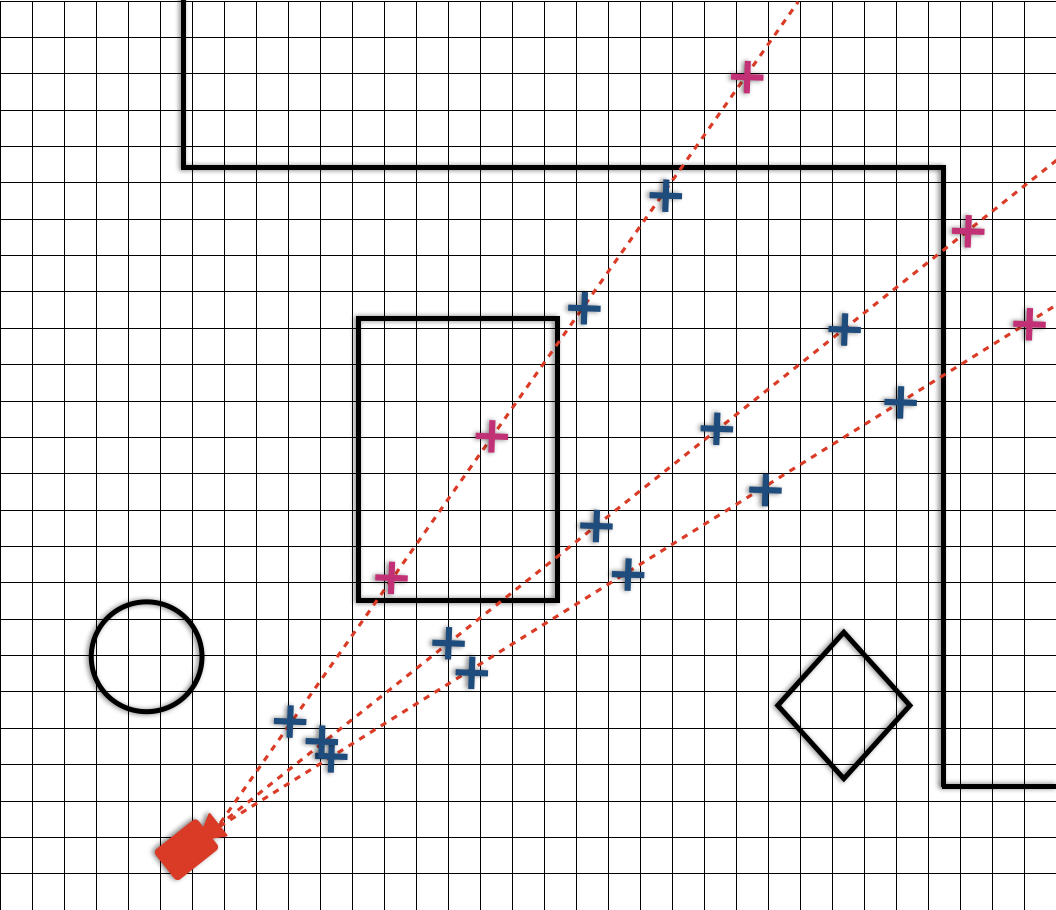} & 
        \includegraphics[width=0.43\textwidth]{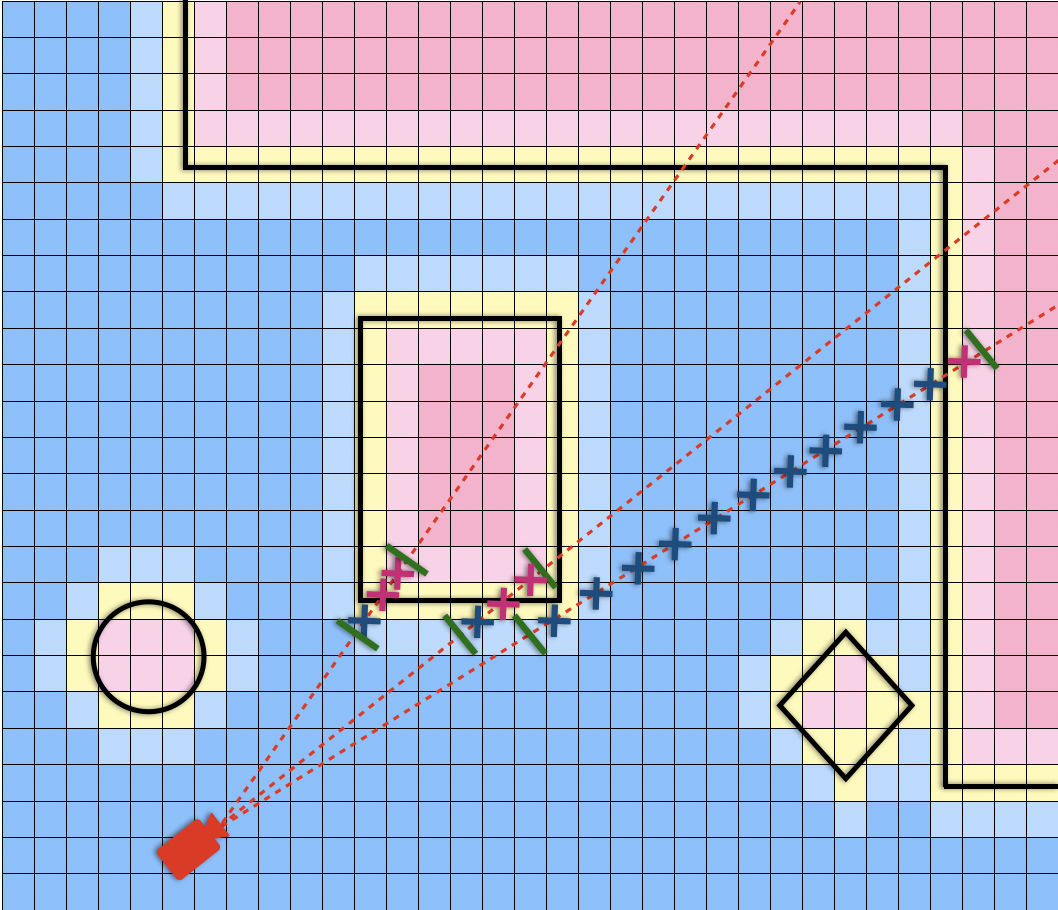}\\
        (a) Inference using Hierarchical sampling &
        (b) Inference using TSDF sampling
    \end{tabular}
    % \caption{Overview of our methodology. Our method is able to automatically decide the near and far plane. In addition, the length between the near and far is various across different rays, by design. We further utilize this variable length in enabling an adaptive the number of samples for each ray, implicitly depending on the characteristics of the scene, which each ray passes through. }
    \caption{
 Three rays and their samples are shown in (a) conventional and (b) ours. The SDF values for the samples (x marks) and grids (in (b)) are colored as red negative and blue positive. 
 Note that both (a) and (b) use an average of 6 samples per ray.
 (b) sets $t_n$ when SDF $\leq 1$ and sets $t_f$ when two consecutive voxels of SDF $\leq 0$ are hit. 
 On the middle ray, the large sample interval ends up passing the inside of the box and causes the first surface to be missed in (a). 
 In (b), however, the sample bound is set correctly and the surface is found.
 On the right ray in (b), if the sampling range were determined only by searching the voxel with SDF $=0$, (b) would erroneously find the periphery of an object as the first face, because the ray passes through a voxel with SDF $=0$ (due to quantization). 
 Our design, which allows to decide the $t_n$ and $t_f$ and to keep on querying the MLPs, can successfully handle the edge cases.
  }
    \label{fig:methodology} 
\end{figure*}

\subsection{Neural Surface Fields}
SDF \cite{sdf} is the implicit representation that encodes geometry in iso-surfaces. Implicit functions like SDF have been reported to define surfaces better than volume density fields \cite{deepsdf,nglod}. VolSDF \cite{volsdf} transforms the SDF value into a volume density using the cumulative distribution function (CDF) of the Laplacian distribution. MonoSDF \cite{monosdf} used the monocular cues obtained from \cite{omnidata} as additional monitoring, achieving high fidelity of geometry. \cite{neus} introduced unbiased re-parametrization of the use of the SDF in differentiable density rendering, and follow-up works \cite{neus2,geoneus,neuralangelo,neuralrecon3dinthewild} are actively presented in the community.  
Our method can be used as a standard module for all previous and future neural surface field models, as long as they use a volume rendering scheme. 
Unlike previous approaches, our method does not require any model modifications. 

\subsection{Sampling Strategies}
Sampling strategy is one of the most important procedures affecting both render quality and computational cost. Sampling at insufficiently dense intervals results in large performance degradation, and unnecessarily large samples increase execution time and memory consumption. For efficient sampling, most schemes \cite{nerf,adanerf,neus,mipnerf, volsdf} sample points hierarchically: coarse and fine sampling. The purpose of coarse sampling is to find out the approximate distribution of rays so that we know where to focus in the next (fine) sampling. \cite{volsdf} proposed Error Bound Sampling, which is a variation of hierarchical sampling and ensures that the error in opacity caused by the discretization does not exceed an upper bound. 
However, its accuracy comes at the cost of more than twice the inference and training time of hierarchical sampling. \cite{donerf, adanerf, mip360,raydf, nerfindetail, nerfacc} have attempted to find better sampling positions by introducing additional networks that need to be trained and queried anew.
\cite{plenoxels, nsvf, ngp, enerf} use 3D grid-based representations to avoid sampling points in empty space. However, the ambiguity of density in deciding surfaces and the discrete nature of grids make them less suitable for highly accurate SDF approximation in large scenes. As a result, geometric accuracy is less explored in previous arts \cite{gaussian_splatting,ngp,enerf} which are mainly designed for skipping spaces of low density.
This motivates us to choose the Neural Surface Field methods as our backbones and also keep the representation continuous. Unlike \cite{donerf,adanerf, mip360, nerfindetail}, we do not introduce any additional neural parameters.

\section{Methodology}
\label{sec:method}

\SetKwInput{kwInput}{Input}
\SetKwInput{kwRequire}{Require}
\SetKwInput{kwOutput}{Output}
\SetKwInput{kwReturn}{Return}
\begin{algorithm}[h]
\caption{TSDF Boundary detection on a ray}

\DontPrintSemicolon
\kwInput{\\
\hspace{1em} $\bold{o}$: camera origin \\
\hspace{1em} $\bold{v}$: ray (unit direction vector) \\
% \hspace{1em} $\mathcal{B}$: NDC \\
\kwRequire{\\
% \hspace{1em} $t_n$: near bound of $t$ to sample\\
\hspace{1em} $t_f$: original far bound of $t$\\
% \hspace{1em} $D_{T}$: truncated distance\\
\hspace{1em} $D_s$: surface criteria\\
\hspace{1em} $M$: maximum number of steps to confirm inside\\
\hspace{1em} $\mathcal{V}$: TSDF volume \\ 
}
\kwOutput{\\
\hspace{1em} $t_n$ : near sample bound\\
\hspace{1em} $t_f$ : far sample bound\\
}\;

% $t_0 \gets  0$\\
% \tcc{1st phase: caving near bound}\\
\tcc{\small Determine near bound $t_n$}
$t \gets 0$\;
$\bold{p} \gets  \bold{o} + t\bold{v}$\;
$X \gets \textsc{Voxel}(\bold{p}$)\;
\While{ $\bold{p} \in \textsc{RenderingCube} $ } { % \in \mathcal{B}
    \If{ $\mathcal{V}(\textsc{VoxelCenter}(X)) \leq D_s$}{
        $t_n \gets t$\;
        \textbf{break}
    }
    $X, t \gets \textsc{NextVoxel}(X, t, \bold{v})$\;
    $\bold{p} \gets  \bold{o} + t\bold{v}$\;
}\;
% \tcc{2nd phase: Carve far bound}
\tcc{\small Determine far bound $t_f$}
$m \gets 0$\;
\While{ $\bold{p} \in \textsc{RenderingCube} $ }{
    \eIf{ $ \textsc{ForAll}(X_{nb} \gets \textsc{NeighborVoxels}(\bold{p}) ;$\\ $\hspace{3.3em} \mathcal{V}(X_{nb}) < 0)$ }{
        $m \gets m+1$
    }{
        $m \gets 0$
    }
    $X, t \gets \textsc{NextVoxel}(X, t, \bold{v})$\;
    $\bold{p} \gets  \bold{o} + t\bold{v}$\;
    \If{ $m == M$ }{
        $t_f \gets t$\;
        \textbf{break}\\
    }
}
}
\kwReturn{
$t_n, t_f$ 
}

\label{alg:spacecarving}
\end{algorithm}

\begin{table*}[!tph]
    % \centering
    \begin{tabular}{c|rrrrr}
        \toprule
        Dataset & Garage + MonoSDF & Lobby + MonoSDF & Replica + NeRF & Replica + MonoSDF \\
        \midrule
        \# of total rays                 & 33M       & 782M  &  14M &  14M \\
        Original Sampling Range (m)  &6.24     & 15.46 &  2.50&  1.35 \\
        Reduced Sampling Range (m)    & 1.15    & 3.80 &  0.21    &  0.08 \\
        \# of failed to find surface  & 134 (0.00003\%) & 26.1k (0.0004\%) &  123k (0.835\%) & 2k (0.014\%)\\
        % fail rate (\%)                & 0.00003   & 0.00041 & 0.00334 & 0.835 \\
        \bottomrule
    \end{tabular}
    \caption{Comparison of the sampling range, determined by our TSDF volume, across different datasets and models. 
    % Original Sampling Range: the average length of the sampling range of the model or dataset. Reduced Sampling Range: the average length of the range achieved by our algorithm. \# number of failed to find surface: the number of rays whose depth used for integration does not fall within our reduced range.
    }
    \label{tab: TSDF integration}
\end{table*}

In this section, we elaborate a novel approach to accelerate the efficiency from the existing sampling methods. We voxelize the implied geometry from the trained model, inspired by the classical TSDF integration. 
We incorporate a geometric prior to define an appropriate interval along each ray in which points are sampled. 
This effectively avoids sampling points that are unnecessary for the rendered results. 
In Sec. \ref{subsection:adaptivesampling}, we introduce our further technique that adapts to different rays with different characteristics. 
% For more efficient rendering, we sample points by uniform gap during initial sampling to prevent missing the first surface because of too few samples.

\subsection{Preliminaries}
{\bf Volume Rendering} 
% Neural representation of the Signed Distance Fields often finds rendering onto the 2D plane useful. 
% This rendering includes color, that the neural field contains by default, and the depth maps.
Volume rendering is one of the most essential parts of the Neural Surface Field. For instance, RGB images and depth maps can be obtained by the volume rendering. 
% NeRF represent a 3D scene as volume density and color field parameterized by fully connected multilayer perceptron (MLP). 
To render the RGB image $\bold{C}$, given the ray $\bold{v}$ starting at the camera origin $\bold{o}$, the rendering algorithm numerically integrates the color radiance $\bold{c}$ at sampled 3D points $\bold{p}(t) = \bold{o}+t\bold{v}$ on the ray for the sample set $\mathcal{T} = \{t_i\}_{i=1}^m$ as follows:
\begin{equation}
\hat{\bold{C}}_\mathcal{T}(\bold{o},\bold{v}) =\sum_{t_i \in \mathcal{T}} T(t_i)\,\alpha(t_i)\,\bold{c}(t_i),  
\label{eq:vol_render}
\end{equation}
where $\alpha(t_i)=1-\text{exp}({-\sigma(t_i)\Delta t_i})$ and $T(t_i)= \prod_{j=1}^{i-1}(1-\alpha(t_j))$ are the opacity and the accumulated transmittance of the i-th ray segment, respectively. 
$(\sigma (t_i),\bold{c}(t_i)) = \bold{MLP}(\bold{p}(t_i),\bold{v})$ denotes the density transformed from SDF value and the RGB color of the point $\bold{p}(t_i)$.

\vspace{.5em} \noindent{\bf Depth} 
Since the accumulated transmittance $T(t)$ decreases as a sample approaches the surface of an object, the weight distribution $w(t)=T(t)\alpha(t)$ ideally appears to be a Dirac $\delta$ function. In other words, the highest weight point corresponds to the first intersection of the ray with an object. 
Using this fact and the eq. (\ref{eq:vol_render}), we compute the depth of a ray as follows:
\begin{equation}
 \hat{D}_{\mathcal{T}}(\bold{o},\bold{v}) = \frac{1}{\sum_{t_i \in \mathcal{T}} w(t_i)} \sum_{t_i \in \mathcal{T}} w(t_i)\, t_i.
\label{eq:compute_depth}
\end{equation}
% \begin{equation}
%  \hat{D}_{\mathcal{T}}(\bold{o},\bold{v}) = \sum_{t_i \in \mathcal{T}}\tilde{w}(t_i) t_i, 
% \label{eq:compute_depth}
% \end{equation}
% where $\tilde{w}(t_i) = w(t_i) / \sum_{t_i \in \mathcal{T}} w(t_i)$ is the normalized weight.\\

% $\bold{Hierarchical \ Sampling.}$ Point sample strategy is one of the most important procedure affecting both render quality and computational cost. Sampling at insufficiently dense intervals results in large performance drops, and unnecessarily large samples increase execution time and memory usage. To efficient sampling, the majority of schemes \cite{nerf,adanerf,neus,volsdf,mipnerf}  hierarchically sample points: coarse and fine sampling. The purpose of the coarse sampling is to find out the approximate distribution of rays so that we know where to focus the next sampling. After calculating the weights $\{ w(t_i) =T(t_i)\alpha(t_i)| t_i \in T_C \}$ of coarse set $\mathcal{T}_C$, fine sampling processes using inverse transform sampling from normalized weights  $\hat{w}(t_i) = w(t_i) / \sum_{t_i \in \mathcal{T}_C} w(t_i)$, which these samples are clustered in locations that have a high impact on rendering.        

\subsection{TSDF Integration}
\label{sec:integration}

To render a novel-view image of the scene, we only need to sample the area near the object surface.
Since the depth for a novel ray is not known, the traditional sampling methods test the entire range to find where the surface is.
Our main intuition is that we can precompute the space occupancy from the trained neural network and store it in a 3D TSDF volume. 
Unlike the color values, it is possible to cache the occupancy for a point because it does not change for different viewing directions.
At render time, the SDF value for a sample can be efficiently queried without inferencing the neural network.

The TSDF integration builds the TSDF value grid $\mathcal{V}$ by ray casting from all images in the training set, similar to the standard TSDF algorithms \cite{tsdf,kinectfusion}.
The surface point $\bold{p}_\bold{v}^*$ for a ray $\bold{v}$ is estimated from the depth of the trained network (Eq. \ref{eq:compute_depth}).
The value of each voxel $\bold{x}$ intersecting with the ray is updated as the weighted average of the clamped SDF value $s_\bold{x}$,
\begin{equation}
  s_\bold{x} = \textsc{CLAMP}(\bold{v} \cdot (\bold{p}_\bold{v}^*-\bold{x}),\;-D_T,\;D_T) ,
\end{equation}
where $D_T$ is a user-defined saturation distance, to reduce the noisy and inaccurate depth estimates from the network. 
We provide the psuedo-code for our TSDF integration algorithm in the supplementary material.

\subsection{Sample Boundary Detection}
\label{subsection:carving}
In this section, we describe our method for minimizing the sampling of empty, unseen, or space inside objects.
To do this,  Algorithm. \ref{alg:spacecarving} determines the sampling bound $(t_n, t_f)$ so that the object surface lies within it, using the pre-computed $\mathcal{V}$ to obtain rough information about the scene geometry. 
To decide $t_n$, we have $\bold{p}_k$ visit the voxels in $\mathcal{V}$ along a ray until it encounters a voxel that falls into the \textit{surface} group, i.e., $s_\bold{X} \leq D_s$.
Then, to determine $t_f$, we march the $\bold{p}_k$ a few more voxels after hitting a $s_\bold{X} < 0$.
It is important to make sure that the $\bold{p}_k$ is actually inside an object. 
To ensure this, we check a certain number of consecutive $\bold{p}_k$'s to see if all their neighboring voxels are negative (occupied).

Fig.~\ref{fig:methodology} illustrates three representative examples of  our motivation for finding $t_n$ and $t_f$. 
For the left ray, ours (b) considers a much smaller range compared to the conventional method \cite{monosdf} (a), and we can use much fewer samples without sacrificing the quality.
The middle ray shows why thin objects can disappear when there are not enough samples in (a), as in Fig. \ref{fig:teaser}.
The last case is when a ray passes very close to the near object but hits the surface of the far object. If $n_f$ is set too close, the bound will not contain the real object surface and will produce an invalid color or depth. Therefore we need to enforce $n_f$ to be definitely inside an object.

There are rare cases where the sampling bound does not contain the true surface, thus producing invalid rendering results. 
We can detect this by checking the TSDF weights in the bound, since the weight sum must be close to 1 if the surface is inside the bound. 
In such cases, we revert to the conventional method of sampling the entire ray range with more samples. 
See the \label{subsection:recovery} section for a detailed discussion.

\subsection{Adaptive Sampling}
\label{subsection:adaptivesampling}
Most existing sampling methods \cite{nerf,mipnerf,volsdf,neus} sample the same number of points in the same depth range over all rays. 
Our algorithm adaptively finds the sampling range per ray according to the scene geometry, and we need to determine how many samples are needed.
If the number of samples for all rays is set uniformly to a small number, the sampling interval for rays with a long range will become too large, resulting in a failure to find the surface at the coarse scale, and it cannot be recovered even with subsequent sampling. 

For more efficient sampling and to eliminate the possibility of missing the surface in coarse sampling, we sample the points at equal intervals within the sampling range along the ray, and obviously the number of samples is proportional to the length of the sampling range. 
Since our sampling ranges are very tight for most rays, our approach can maintain rendering quality with a much smaller average number of samples than conventional methods.

\section{Experimental Results}
\label{sec:exp}

\begin{figure*}[!t]
    \renewcommand{\tabcolsep}{1.0pt}
    \centering \small
    \begin{tabular}{ccc}
    \includegraphics[width=0.35\textwidth, height=0.3\textwidth]{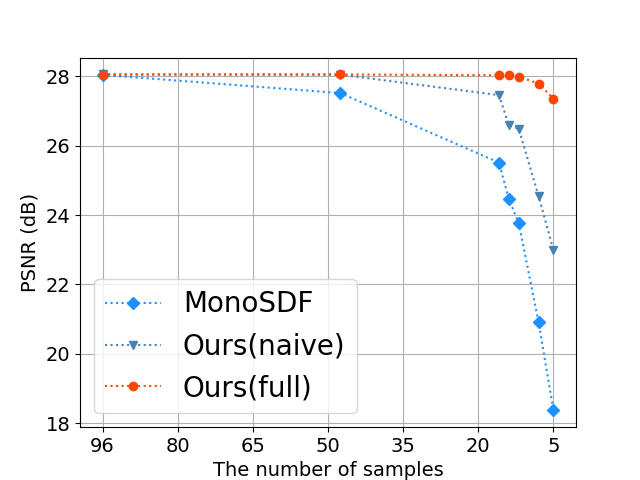} &
     \includegraphics[width=0.35\textwidth,height=0.3\textwidth]{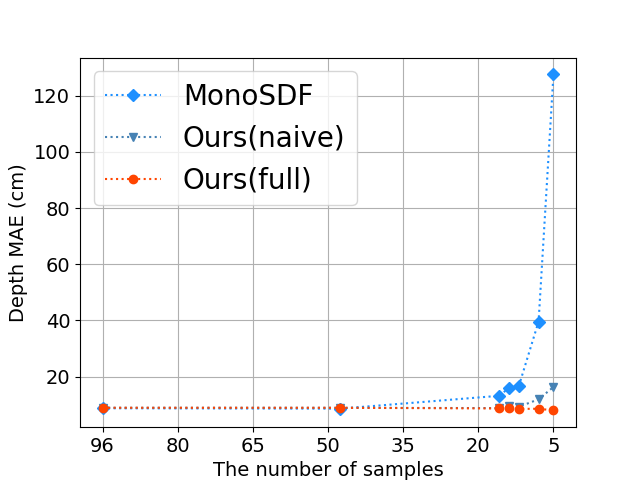} &
     \includegraphics[width=0.35\textwidth,height=0.3\textwidth]{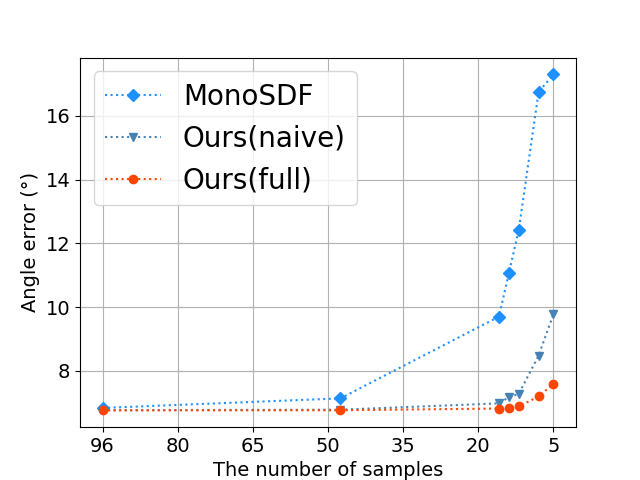} \\
     (a) PSNR & (b) Depth MAE & (c) Angle error
    \end{tabular}
    \caption{Ablation study on Garage by the number of samples. From left to right: PSNR, depth mean square error (MAE), and surface normal angular error.
    % Across all metrics, \cite{monosdf} shows the largest drop, especially in depth MAE. 
    % Our naive method has relatively small error in depth, compared to other metrics, because the range over which depth is calculated is enforced by our method.
    }
    \label{fig:samples_vs_performance}
\end{figure*}

We evaluate our proposed method using MonoSDF on a real large scene (Lobby), a synthetic scene (Garage) and a public dataset (Replica \cite{replica}). \cite{monosdf} has several encoding options, including single resolution, MLP, and hash encoding from \cite{ngp}, in which we chose the multi-resolution hash encoding for its overall superior quality and efficiency shown in \cite{monosdf}. Additionally, we apply our method to NeRF \cite{nerf} on Replica to verify the effectiveness when the network has poor geometry information. The training progress for each dataset is described in the supplementary material. 

\subsection{Datasets}
{\bf Lobby} This is a real-world dataset 
 that was captured using a mobile robot with four ultra-wide field-of-view (FOV) fisheye cameras in an indoor scene. We applied \cite{omnislam} to the data to obtain the camera poses in the scene. {\bf Lobby} contains 1552 images with a resolution of $1344 \times 1080$ and a FOV of 220 degrees. 
We limit the FOV to $\ang{190}$ to address the vignetting problem during training. 
To mimic the depth and normal prior in \cite{monosdf}, we estimate the depth maps from the images using \cite{omnimvs} and compute the surface normal map by backprojecting the 2D points using depth and performing the cross product between the nearest 3D points.

{\bf Garage} 
This dataset was created with Blender from a 3D model of a garage with various objects. 
The camera setup is the same as in Lobby. The Garage dataset consists of 80 images of size $832 \times 832$ and their corresponding depth and normal maps. We use this dataset to evaluate the geometric performance as it can provide accurate depth and normal.

{\bf Replica}
We use the officially provided Replica \cite{replica} dataset from MonoSDF \cite{monosdf}, which contains the monocular depth and normal maps from Omnidata \cite{omnidata}.
We choose room 0, which consists of 100 images of size $384 \times 384$.  
NeRF \cite{nerf} is trained on RGB images only, but only for the quantitative comparison of the TSDF volume in Tab. \ref{tab: TSDF integration}, we trained \cite{monosdf} with the monocular cues.           

\subsection{Implementation Details}
\label{seq:tsdf_parameter}
In all our experiments, the TSDF grid resolution is set to $512^3$.
When the pre-trained model is MonoSDF \cite{monosdf}, we set the maximum distance $D_T$ and $D_s$ to $5 \times$ and $1 \times$ the voxel size, respectively. 
To determine $t_f$ in the TSDF Boundary Carving process, we explore all $5\times5\times5$ voxels around the voxel of interest and test if they are inside an object. 
In the case of NeRF \cite{nerf}, due to the low quality of its density field, we set $D_T$ and $D_s$ to $39 \times$ and $27 \times$ of the voxel size and explore the $7\times7\times7$ neighboring voxels. 
To ensure that the voxel of interest is really inside an object, we break the second while loop in the algorithm \ref{alg:spacecarving} only when such a neighborhood criterion is satisfied 15 times in a row. 
For comparison, the standard hierarchical sampling in state-of-the-art methods is used, i.e. MonoSDF\cite{monosdf}. 

\subsection{TSDF Volume Evaluation}
To integrate a TSDF volume, the algorithm \ref{alg:integration} takes 27.02 seconds for modelling the entire Garage scene. Once the TSDF volume is integrated, the TSDF volume can serve as a strong geometric prior to provide sampling bounds in infinite number of incoming novel views. If voxels which has not been updated by the rays are found, we can cast additional rays to fill them up, until sufficiently large space is covered. 
%27.021233558654785s for garage (when using batch size of 32k rays). 39046464 rays in total. 
%0.692$\mu s$ to update TSDF voxels along a ray

Tab. \ref{tab: TSDF integration} describes the extent to which the sampling range has been reduced for each dataset. For all datasets, the average sampling range was reduced by $\bold{24\%}$ or less compared to the original range in rays of the train set. 
% Note that the integration of the TSDF volume takes only 00 seconds to model a  $\text{m}^3$ scene. 
Then, the TSDF volume can be used to provide a strong guidance for numerous upcoming novel views. We analyze the cases where the depth used for integration does not fall within the reduced sampling range of our method.

This rare exception can occur due to the discrete nature of the grid representation and the finite number of samples on a ray. 
For example, the Garage using MonoSDF shows very few errors when trained with accurate depth and normal. 
However, if the pre-trained model has unstable geometry, such as NeRF, the failure rate can increase significantly, but is still below 1\%.
When integrated with depth obtained from NeRF without any supervision except for color, the failure rate comes at 0.84\%, which is about 60 times higher than that of the same Replica \cite{replica} with MonoSDF (the last column).

To deal with these cases we introduce the recovery method to our algorithm as in Sec. \ref{subsection:recovery}. This allows us to detect the rays where the true surface is outside the sampling range, and only for these rays, we increase the sampling range to the entire ray length, so that we can efficiently avoid such failure cases.

% \begin{figure*}[!tp]
% \centering
% \includegraphics[width=0.5\textwidth, height=0.4\textwidth]{figures/Garage_psnr_sample.png}
% \caption{Plots for Garage. Same plot will be available for other datasets as well.}
% \label{fig:sample_psnr}
% \end{figure*}

\subsection{Performance}
In Tab. \ref{tab:lobby}, we report the rendering time, PSNR, and SSIM \cite{ssim} in the Lobby dataset, with respect to the sampling methods and the number of samples. When the number of samples is sufficiently large (96), the performance  is almost the same for all metrics, regardless of the sampling method. However, when the number of samples is limited to 12, the PSNR of \cite{monosdf} and our method deviate significantly to 18.42 and 19.77, respectively, indicating the advantages of our method.

\begin{figure*}[!tp]
    \renewcommand{\tabcolsep}{1.1pt}
    \centering \footnotesize
    \begin{tabular}{ccccccc}
        \includegraphics[width=0.13\textwidth]{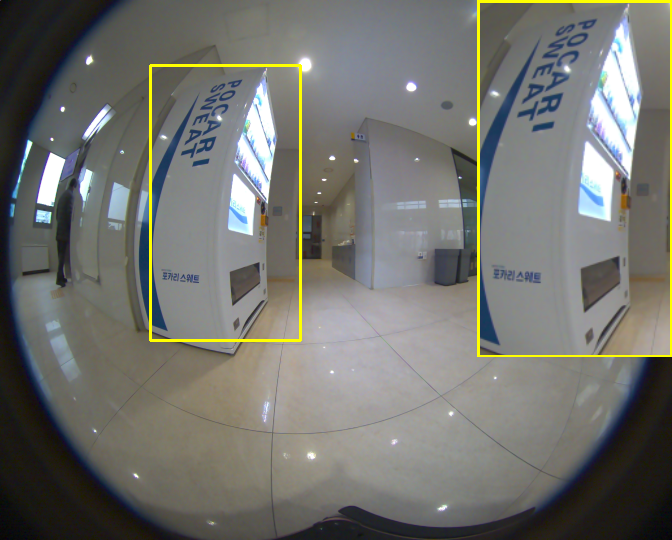} &
        \includegraphics[width=0.13\textwidth]{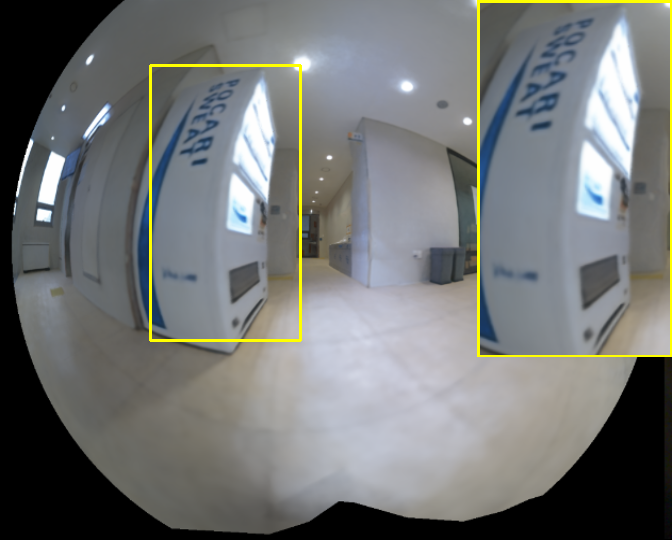} &
        \includegraphics[width=0.13\textwidth]{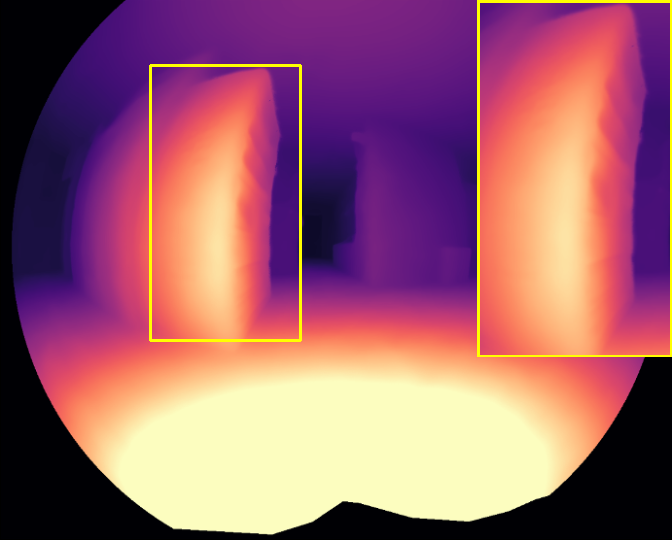} &
        \includegraphics[width=0.13\textwidth]{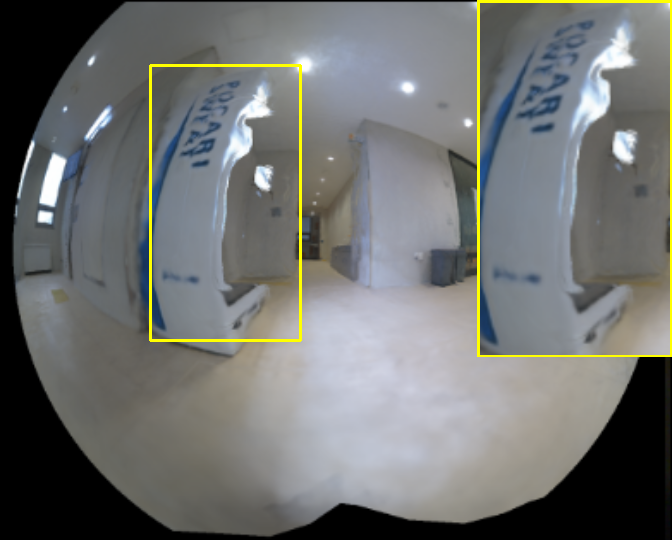} &
        \includegraphics[width=0.13\textwidth]{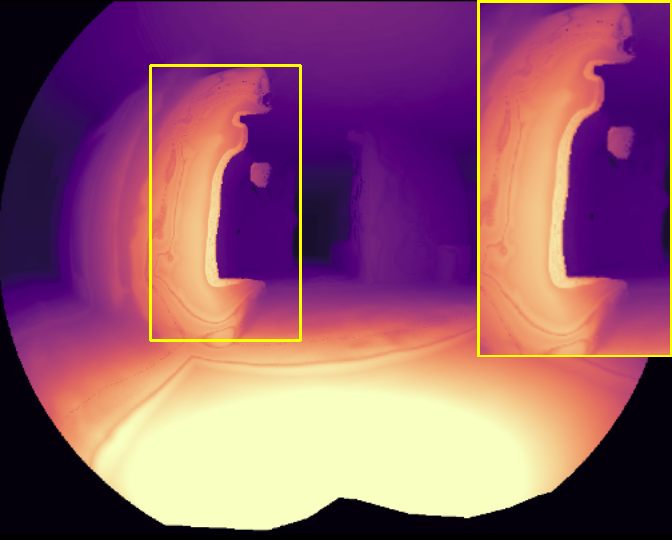} &
        \includegraphics[width=0.13\textwidth]{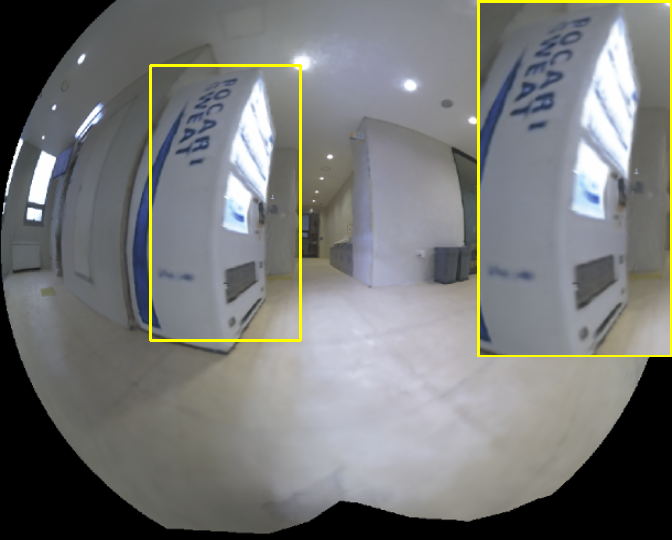} &
        \includegraphics[width=0.13\textwidth]{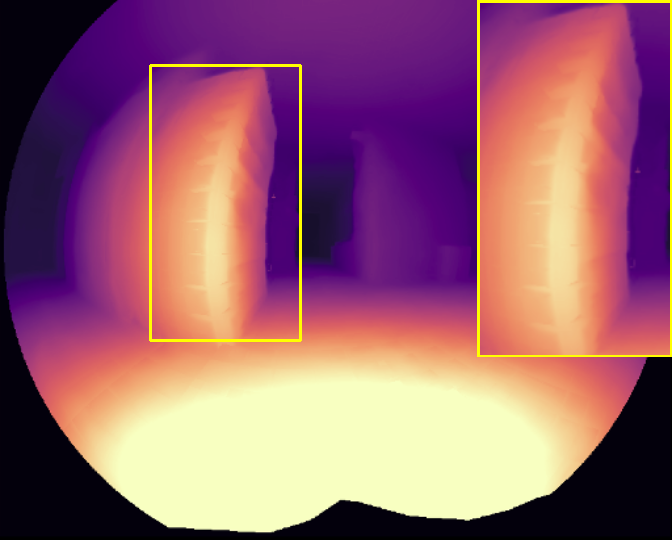} \\
        \includegraphics[width=0.135\textwidth]{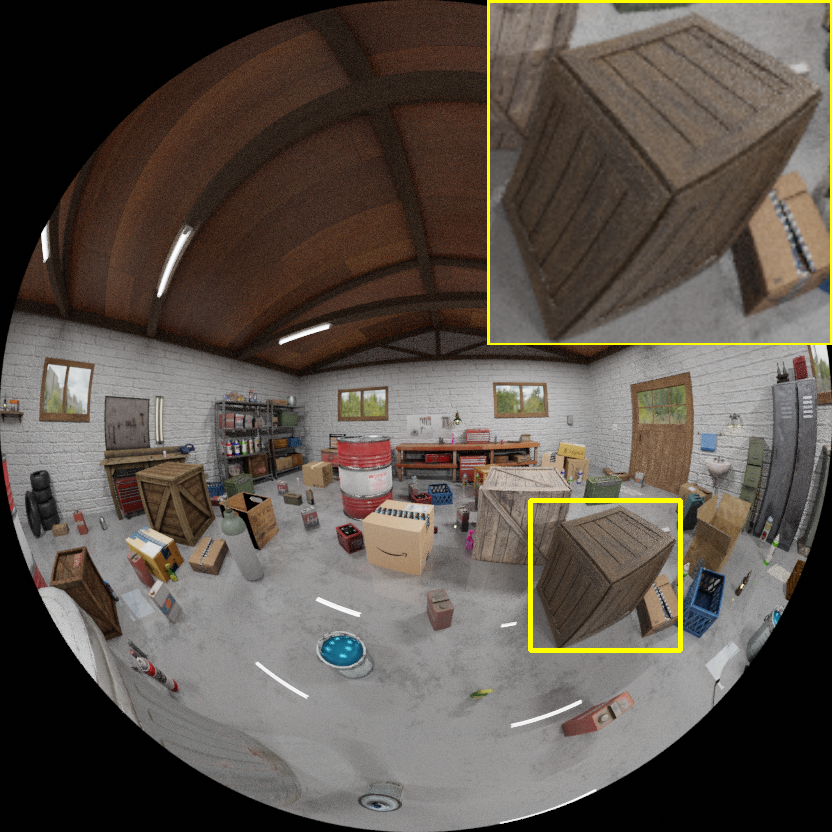} &
        \includegraphics[width=0.13\textwidth]{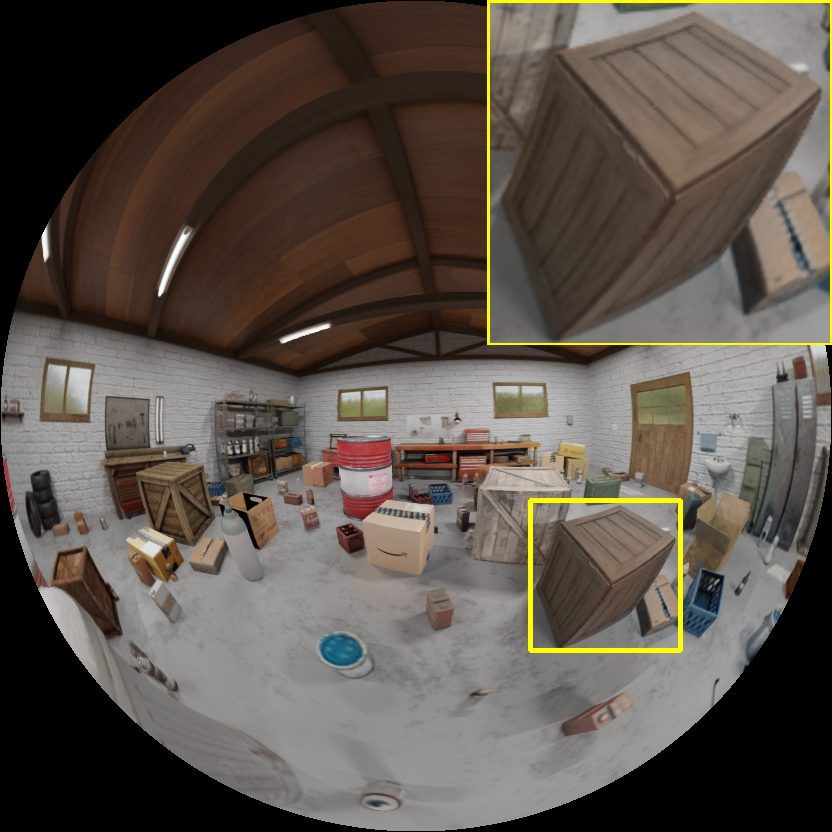} &
        \includegraphics[width=0.13\textwidth]{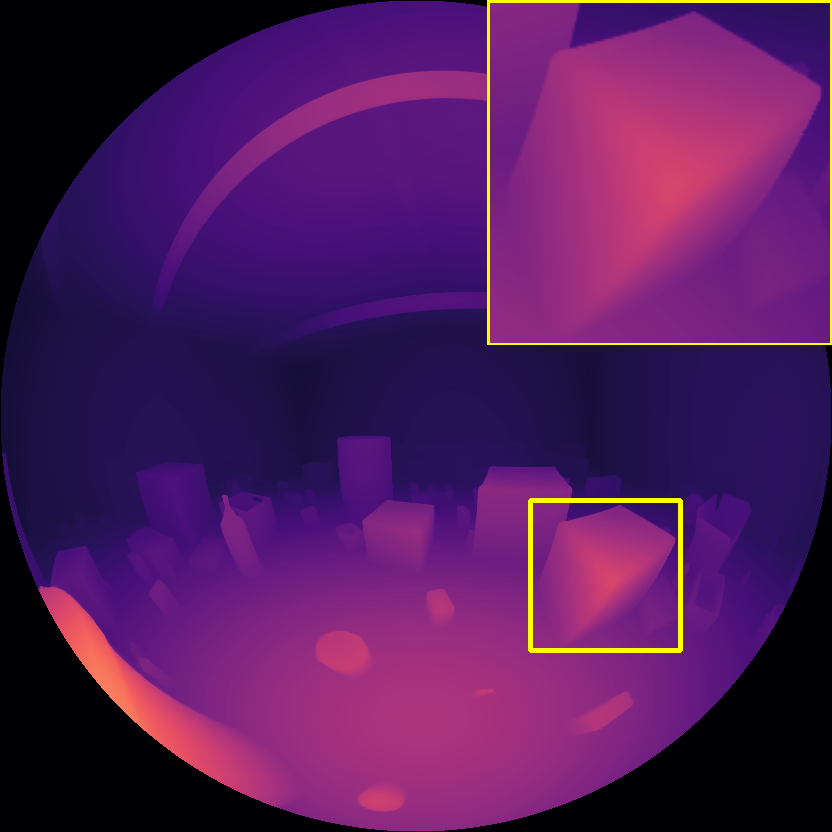} &
        \includegraphics[width=0.13\textwidth]{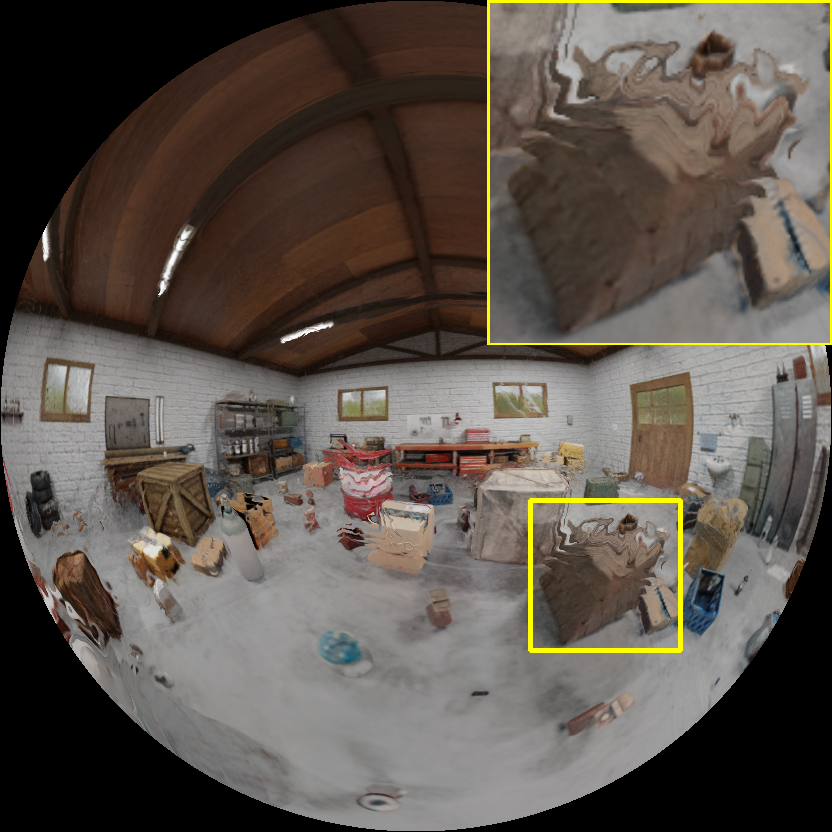} &
        \includegraphics[width=0.13\textwidth]{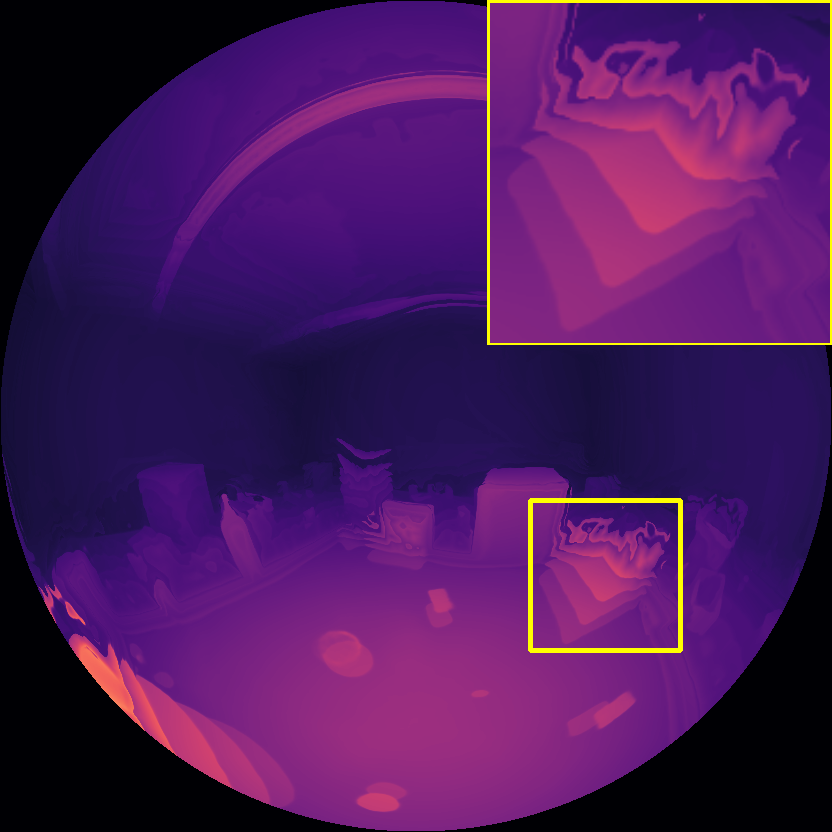} &
        \includegraphics[width=0.13\textwidth]{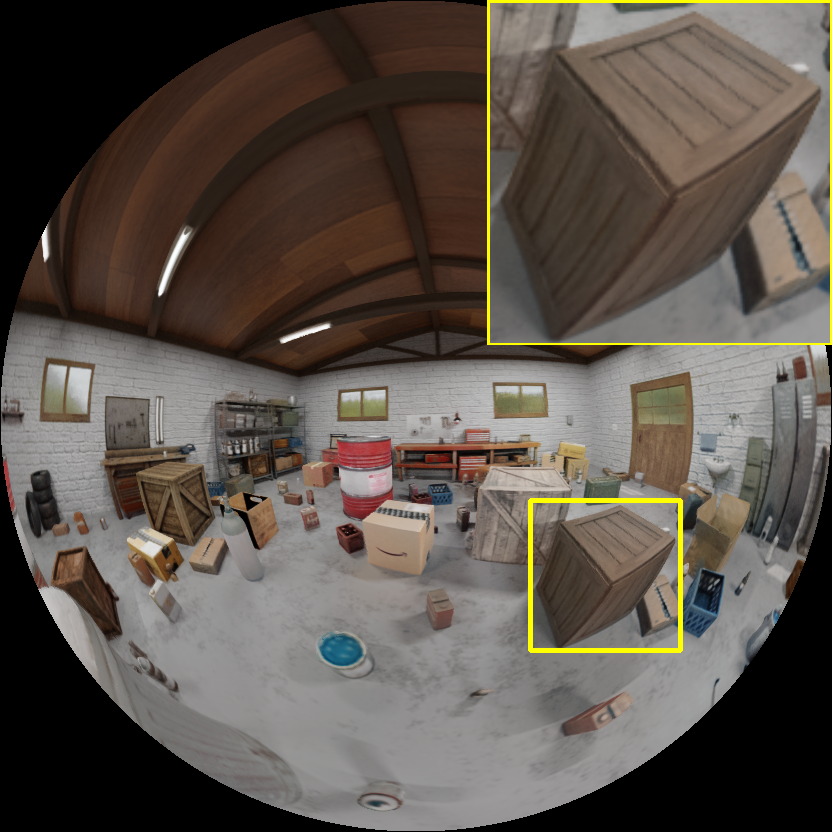} &
        \includegraphics[width=0.13\textwidth]{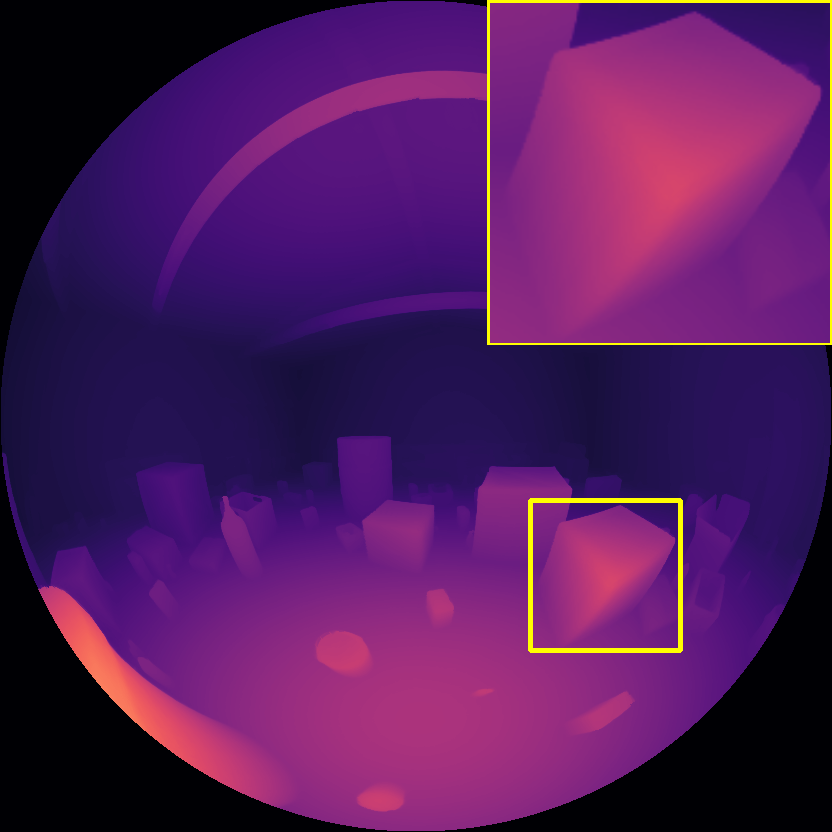} \\
        \includegraphics[width=0.13\textwidth]{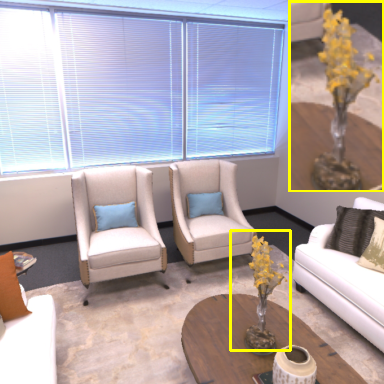} &
        \includegraphics[width=0.13\textwidth]{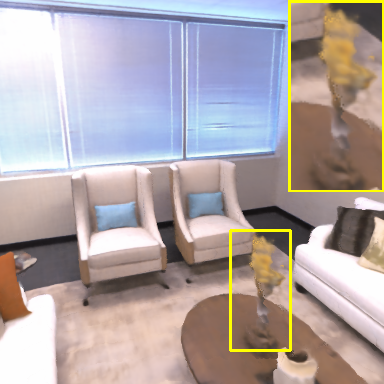} &
        \includegraphics[width=0.13\textwidth]{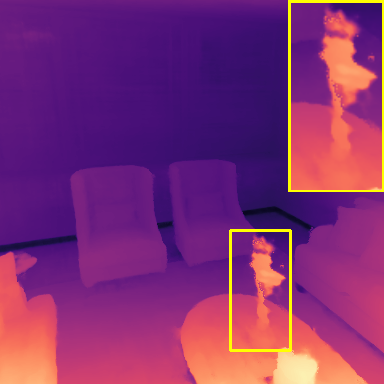} &
        \includegraphics[width=0.13\textwidth]{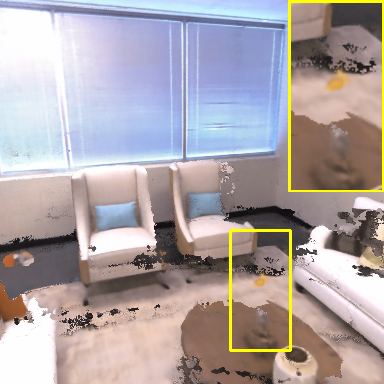} &
        \includegraphics[width=0.13\textwidth]{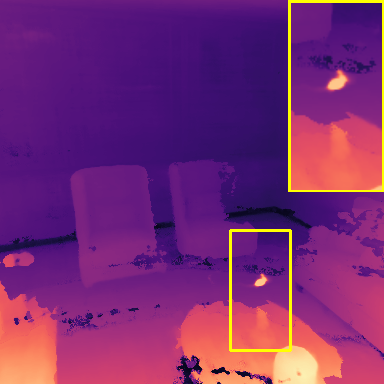} &
        \includegraphics[width=0.13\textwidth]{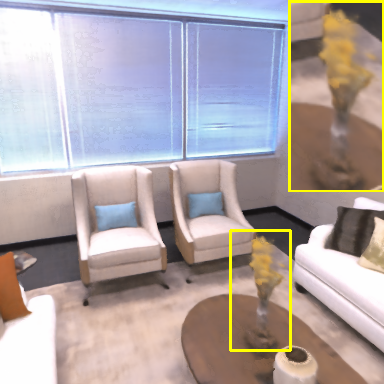} &
        \includegraphics[width=0.13\textwidth]{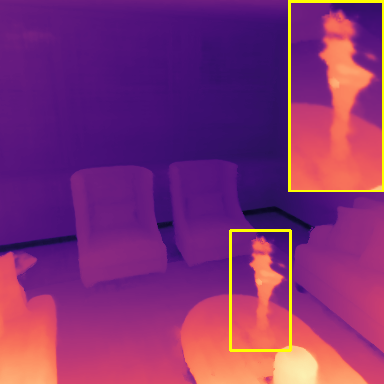} \\
        (a)   & (b) & (c)  & (d)  & (e)   & (f)  & (g)   \\
    \end{tabular}
    \caption{Qualitative results on \textbf{Lobby}, \textbf{Garage}, and \textbf{Replica} \cite{replica} from top row to bottom. (a): ground truths. (b), (c): base models \cite{nerf,monosdf} results with large number of samples. (d), (e): base  models \cite{nerf,monosdf} results with small number of samples , and (f), (g): Ours (full) results with small number of samples. 
    % On a real large scene, a synthetic scene, and a public dataset, our method robustly finds the surface, and thus renders color and depth map in higher quality, in reduced number of samples
    }
    \label{fig:qualitative}
\end{figure*}

In Fig. \ref{fig:qualitative} (d), we demonstrate that simply reducing the number of samples causes the invisible area behind the near object to appear. 
Fig. \ref{fig:raydist} effectively shows that this is because the locations of the coarse samples (cyan) are too far apart, causing the first surface to be missing, which in turn causes the surface in the subsequent samples (magenta) to be missing. On the other hand, in (c), our method can find the surface with the same number of samples, because the course samples are distributed within our proposed sampling range.

We evaluate two additional metrics (depth mean square error and normal angle error) on Garage to show that we can preserve not only the photometric quality but also the geometric performance. 
Tab. \ref{tab:garage} shows that as the number of samples decreases, our method performs remarkably on par with results from much larger numbers of samples, with respect to depth and normal.
However, \cite{monosdf} incurs a notable performance overhead of about twice the increase in both depth and normal errors due to the missing first surfaces.

\subsection{Ablation Study}
We compare the performance of the naive method and that of our proposed method, including the adaptive sampling ranges and the variable number of samples on each ray, according to the length of the sampling range. The naive method uses the same number of samples for each ray. 
Fig. \ref{fig:samples_vs_performance} shows the stark differences in performance when the number of samples is reduced from 96 to 5. When the number of samples is halved, our naive method still maintains rendering quality. However, when the number of samples is reduced to less than 16, the naive method starts to lose performance, while our full method is much less affected. The detailed qualitative results are available in the Supplementary Materials.

\begin{table*}[!htbp]
    \centering
    \renewcommand{\arraystretch}{1.1}
    \scalebox{0.9}{
    \begin{tabular}{l|ccrrcrrr}
        \toprule
          Approaches & Sampling &Samples &$\Delta t$ [cm]& Time [s]$\downarrow$  & PSNR [dB] $\uparrow$  & SSIM $\uparrow$ & D [cm]$\downarrow$  & N [deg]$\downarrow$  \\
        \midrule
         
            Mip-NeRF \cite{mipnerf} &HS& 64+32 &17.2& 28.00 & 27.35 & 0.984 & 8.42 & 32.2\\
            % Instant-NGP\cite{ngp}  &RM& 31 &-& 3.30   & 27.84 & 0.977 & 8.64 & 27.5\\
            MonoSDF \cite{monosdf}  &EB& 64+32 &17.2& 54.76 & 26.90 & 0.979 & 9.36 & 6.73\\
            \hline
            MonoSDF \cite{monosdf}        &HS& 64+32&17.2 & 19.62 & 28.05& \textcolor{silver}{0.980}& \textbf{\textcolor{gold}{8.72}}& 6.84\\
            Ours (naive)             &TSDF& 64+32&0.9 & 18.06 & \textbf{\textcolor{gold}{28.06}}& \textbf{\textcolor{gold}{0.981}}& \textcolor{silver}{8.89}& \textcolor{silver}{6.77}\\
            Ours (full)              &TSDF& 64+32 &0.9&18.27 & \textcolor{silver}{28.06}& \textcolor{silver}{0.980} & 8.90 &\textbf{\textcolor{gold}{ 6.76}} \\
            % \hline
      %       MonoSDF\cite{neus}        &HS& 48 &91.95& 10.784&27.520&0.9779&8.583&7.140\\
      %       Ours (naive)             &TSDF& 48 &3.281& 11.113&28.052&0.9804&8.888&6.785\\
      %       Ours (full)              &TSDF& 48 &3.281& 10.775&28.057&0.9805&8.887&6.767\\
      % \hline
      % MonoSDF\cite{neus}         &HS& 16 &137.9& 6.351&25.515&0.9648&13.157&9.703\\
      %       Ours (naive)             &TSDF&16& 7.013& 6.307&27.461&0.9776&8.786&6.983 \\
      % Ours (full)              &TSDF& 16 & 7.013&5.680&28.034&0.9804&8.657&6.822\\ 
      \hline
            MonoSDF \cite{monosdf}        &HS& 6+8& 183.9&  6.10 & 24.46 & 0.955& 15.87 & 11.07\\
            Ours (naive)             &TSDF& 6+8 &10.4& 6.12 & \textcolor{silver}{26.60}& \textcolor{silver}{0.973}&\textcolor{silver}{9.50}&\textcolor{silver}{7.18}\\
            Ours (full)             &TSDF& 6+8 &10.4& 5.26&\textbf{\textcolor{gold}{28.04}}&\textbf{\textcolor{gold}{0.980}}&\textbf{\textcolor{gold}{8.65}}&\textbf{\textcolor{gold}{6.85}}\\
            \hline
            MonoSDF \cite{monosdf}        &HS& 6+6 &183.9&  5.96&23.77&0.947&16.57&12.42 \\
            Ours (naive)             &TSDF& 6+6&10.4 & 5.97&\textcolor{silver}{26.49}&\textcolor{silver}{0.972}&\textcolor{silver}{9.23}&\textcolor{silver}{7.28} \\
            Ours (full)              &TSDF& 6+6 &10.4& 5.05&\textbf{\textcolor{gold}{28.00}}&\textbf{\textcolor{gold}{0.980}}&\textbf{\textcolor{gold}{8.50}}&\textbf{\textcolor{gold}{6.89}} \\ 
            % \hline
            % MonoSDF\cite{neus}        &HS& 8 &275.85& 5.391&20.9029&0.89743&\textcolor{silver}{19.609}&16.751\\
        % Ours (naive)             &TSDF& 8 & 16.9&5.483&\textcolor{silver}{124.5472}&\textcolor{silver}{10.9560}&11.913&\textcolor{silver}{18.482}\\
        %     Ours (full)              &TSDF& 8&16.9 & 4.514&\textbf{\textcolor{gold}{27.7952}}&\textbf{\textcolor{gold}{0.9792}}&\textbf{\textcolor{gold}{8.551}}&\textbf{\textcolor{gold}{7.200}}\\ 
        % \hline
        %MonoSDF\cite{neus}        &HS& 5 &367.8 &4.905&18.366&0.8356&127.6&17.303\\
        %     Ours (naive)             &TSDF& 5 &17& 5.047&\textcolor{silver}{22.993}&\textcolor{silver}{0.9367}&\textcolor{silver}{16.298}&\textcolor{silver}{9.785}\\
        %     Ours (full)              &TSDF& 5 &17& 4.203&\textbf{\textcolor{gold}{27.3608}}&\textbf{\textcolor{gold}{0.977}}1&\textbf{\textcolor{gold}{7.9612}}&\textbf{\textcolor{gold}{7.598}}\\
        \bottomrule
    \end{tabular}
    }
\caption{Quantitative comparison on the \textbf{Garage} dataset. HS and EB denote Hierarchical sampling\cite{nerf,nerfinthewild,mipnerf,neus} and the Error bound sampling\cite{monosdf,volsdf}, respectively. The numbers of coarse and fine samples are combined with $+$. D[cm] is the depth error, measured by MAE. N[deg] is the normal error in degree.  
    % The average number of samples of Ours (full) method has been rounded off to integer for simplicity
    $\Delta t$ refers to the interval between coarse samples. 
    With 96 samplings per ray, the performance between \cite{monosdf} and ours are not significant. However, when the limited number of samples are used to boost the neural SDF representation, ours outperforms \cite{monosdf} in a significant amount, especially in geometry, i.e., depth and normal. 
    % For the page limitation, we provide more results on the Supplementary Material.
    }
    \label{tab:garage}
\end{table*}

\begin{table*}[!htbp]
    \centering
    \scalebox{0.9}{
    \begin{tabular}{c|ccrrcr}
        \toprule
         Approaches&Sampling&Samples& $\Delta t$ [cm]& Time [s]$\downarrow$& PSNR [dB] $\uparrow$& SSIM $\uparrow$\\
        \midrule
              % Instant-NGP\cite{ngp} &RM&243 &8.502  & 21.34 & 16.84 & 0.7438 \\
              MonoSDF \cite{monosdf} &EB&64+32  &21.52   & 102.20 & 19.99 & 0.851 \\
              \hline
              MonoSDF \cite{monosdf}       &HS&64+32  &21.52   & 59.39 &\textbf{\textcolor{gold}{20.01}} & \textbf{\textcolor{gold}{0.848}} \\
              Ours (naive)     &TSDF&64+32  &1.34   & 58.37 &\textcolor{silver}{20.00}&\textcolor{silver}{0.847}\\
              Ours (full)                 &TSDF&64+32  &1.34   & 44.81 & 19.99 & 0.847 \\
              \hline
              MonoSDF \cite{monosdf}       &HS&6+8  &172.10  & 14.57 & 18.56 & \textcolor{silver}{0.759} \\
              Ours (naive)      &TSDF&6+8  &10.70   & 14.28 &\textcolor{silver}{ 18.61} & 0.759 \\
             Ours  (full)                &TSDF&6+8  &10.70   & 12.97 & \textbf{\textcolor{gold}{19.93}} &\textbf{\textcolor{gold}{ 0.845}} \\
             \hline
              MonoSDF \cite{monosdf}       &HS&6+6  &172.10  & 13.30 & 18.42 & 0.753 \\
              Ours (naive)     &TSDF&6+6  &10.70    & 13.59 &\textcolor{silver}{18.58}& \textcolor{silver}{0.762} \\
              Ours   (full)               &TSDF&6+6  &10.70   & 12.91 & \textbf{\textcolor{gold}{19.77}} & \textbf{\textcolor{gold}{0.838}} \\

        \bottomrule
    \end{tabular}
    }
    \caption{Quantitative results on the \textbf{Lobby} dataset. The notations are same with those of Tab. \ref{tab:garage}. As the number of average samples increases, the improvement of our model upon \cite{monosdf} becomes significant. }
    \label{tab:lobby}
\end{table*}

 \begin{table}[!tp]
    \centering
    \scalebox{0.75}{
    \begin{tabular}{c|cccccc}
        \toprule
        Approaches & Samples & Recovery &Time [s] $\downarrow$ & PSNR [dB] $\uparrow$  & SSIM $\uparrow$ & \\
        \midrule
            NeRF \cite{nerf}   & 64+128 & -    & 6.83    & 34.05 & 0.995  &  \\
            NeRF \cite{nerf}   & 32+64 & -    & 2.56    & 28.04 & 0.980  &  \\
            \hline
            NeRF \cite{nerf}   & 16+32  & -    &1.31      & 21.91 & 0.923  & \\
            Ours (naive) & 16+32  &  -    & 1.74   & 27.57 & 0.979 & \\
            Ours  (full)    & 16+32  &  -    & 1.24 & 27.57 & 0.980 & \\
            Ours (naive) & 48+0  & \checkmark   & 2.60 & \textcolor{silver}{33.41}& \textcolor{silver}{0.994}& \\
            Ours  (full)   & 48+0  & \checkmark   & 2.42 & \textbf{\textcolor{gold}{33.65}} & \textbf{\textcolor{gold}{0.995}} &  \\
        \bottomrule
    \end{tabular}
    }
    \caption{Quantitative results of recovery algorithm on Replica \cite{replica}. The sample column is the average number of samples before applying recovery(initial sample + fine samples).}
    \label{tab:replica}
\end{table}

\subsection{Recovery from Incorrect TSDF Estimates}
\label{subsection:recovery}
We further investigate the ability of our method to work with a pre-trained model with less accurate geometry, such as NeRF\cite{nerf}, which was designed as a neural radiance field and thus has noisy geometry.
Although NeRF does not explicitly handle SDF, our method can be applied to NeRF since our algorithm requires only the occupancy estimates from the model and can serve as a modular building block for a variety of volume rendering models. 

\begin{figure}[!tp]
    \renewcommand{\tabcolsep}{2.0pt}
    \footnotesize \centering
    \begin{tabular}{cc}
        \includegraphics[width=0.22\textwidth]{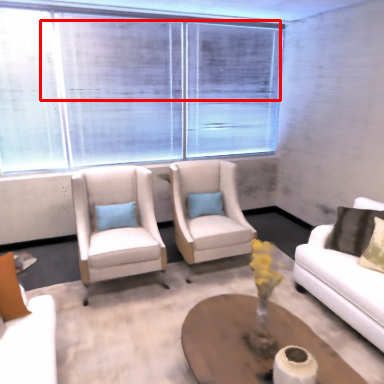} &
        \includegraphics[width=0.22\textwidth]{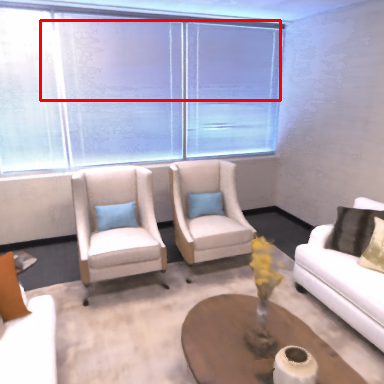} \\
        ours w/o recovery  & ours  \\
    \end{tabular}
    \vspace{-1em}
    \caption{Rendering results with and without recovery. Note that the thin blinds are not correctly rendered without recovery process.
    % What appears to be black inside the red box, especially in extremely fine structures, is due to the sample range being too narrow, causing the overall weight to be too small and the color value to be close to zero.
    }
    \label{fig:recovery}
\end{figure}

In Fig. \ref{fig:recovery}, the challenging fine structures show vulnerability when the reduced sampling range is applied naively. Given these challenges, to increase the flexibility of our method, we present another modification of our method: the recovery algorithm. 

If the sum of the weights in a ray turns out to be less than a threshold, we recover the color of the ray by resetting the sampling range to the initial range and reconducting the sampling.  
It turns out that only 11\% of the total rays trigger recovery on NeRF with reduced samplings, and the result is on par with \cite{nerf}, while cutting the inference time in 30\% than \cite{nerf}.  The outstanding differences occur on the challenging fine structures in Fig. \ref{fig:recovery}, where our recovery algorithm effectively handles such cases where the reduced sample bound does not include the true surface.

\begin{figure*}[!tp]
    \renewcommand{\tabcolsep}{2.5pt}
    \centering \small
    \begin{tabular}{ccc}

     \includegraphics[height=0.1458\textwidth]{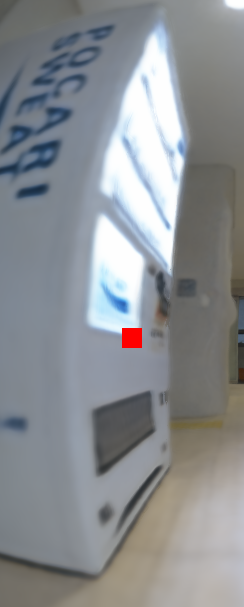} &
    \includegraphics[height=0.1458\textwidth]{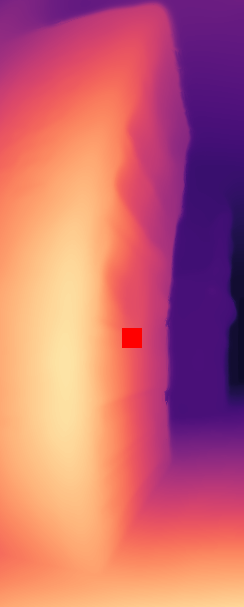} &
    \includegraphics[width=0.7\textwidth,height=0.1458\textwidth]{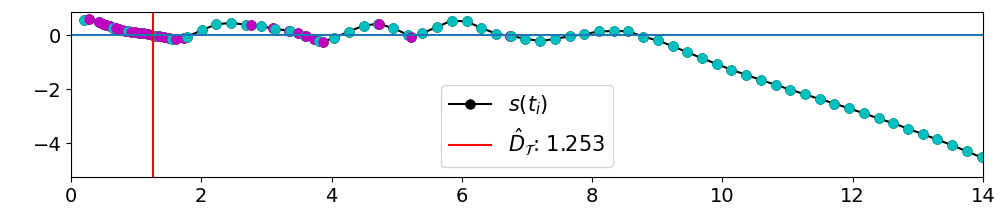} \\  % 0.1458 0.1250
    &&(a) MonoSDF\cite{monosdf}, 96 samples\\
    \includegraphics[height=0.1458\textwidth]{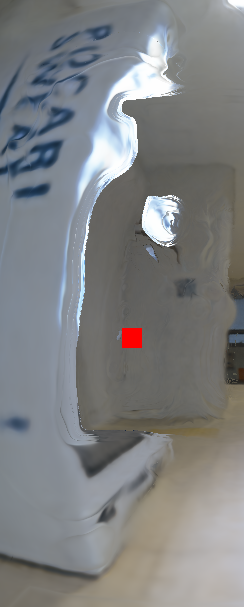} &
    \includegraphics[height=0.1458\textwidth]{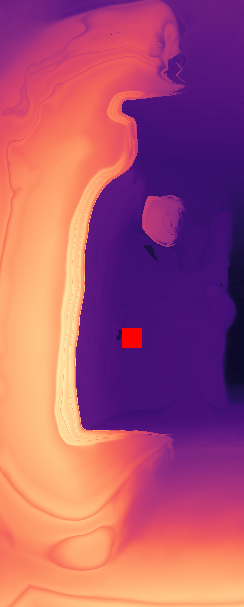} &
    \includegraphics[width=0.7\textwidth,height=0.1458\textwidth]{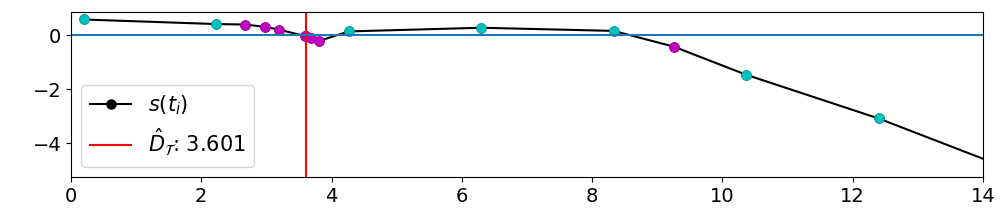} \\
    &&(b) MonoSDF\cite{monosdf}, 16 samples\\
     \includegraphics[height=0.1458\textwidth]{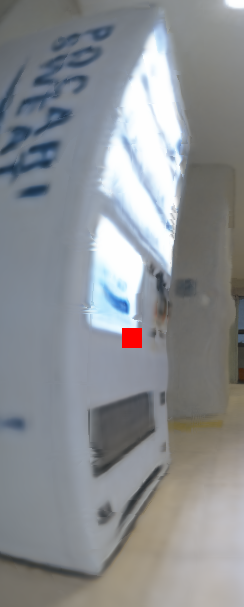} & 
     \includegraphics[height=0.1458\textwidth]{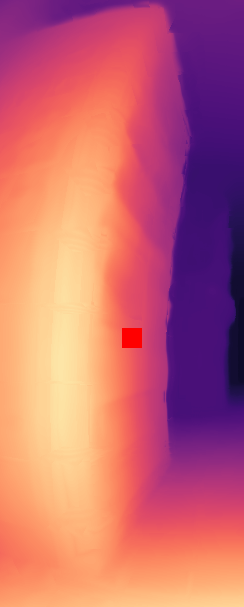}
     &\includegraphics[width=0.7\textwidth,height=0.1458\textwidth]{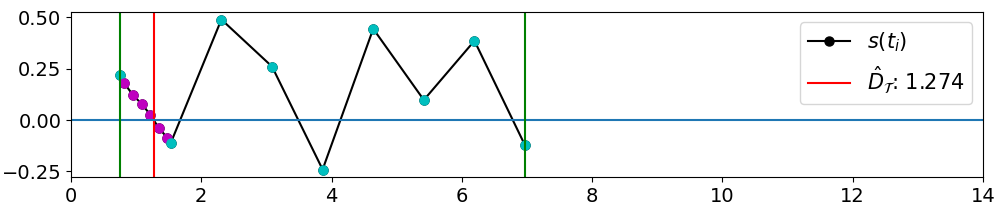} \\
     &&(c) Ours, 16 samples\\
     % (a) MonoSDF\cite{monosdf} with 96 samples & (b) MonoSDF\cite{monosdf} with 16 samples & (c) Ours (full) with 16 samples
    \end{tabular}
    \caption{The cyan and magenta markers indicate coarse and fine samples, respectively. With as many as 96 sample, many meaningless samples are queried in (a). Given 8 coarse samples respectively, (b) illustrates how a limited number of coarse samples may lead to skip the first surface, while our method can detect the surface in (c), since we bound the range (green vertical lines) by exploiting the model-inherent TSDF prior. 
    % More extensive analysis on easier and harder cases are available in Supplementary Material. 
    % Medium-hard case is shown. We also conducted analysis on  easy and hard case, respectively, provided in the Supplementary Material.  We illustrate coarse sample points in magenta markers and fine sample points in cyan markers along a ray in the red marker on the result image. }
    % shows that our method can more easily solve it
    }
    \label{fig:raydist}
\end{figure*}

\section{Conclusion}
We present a novel approach that significantly reduces the number of samples while maintaining the rendering quality, using the traditional TSDF volume. Given a pre-trained model, we construct a single resolution TSDF voxel grid using the ray termination depth. For each ray, we compute the sampling bound, which encompasses the first surface the ray encounters. 
In addition, we adaptively adjust the number of samples according to our sampling bound. 
Given that the choice of near and far planes has been considered a very sensitive hyperparameter for neural field researchers, the ability of our method to automatically decide the near and far range in a ray-adaptive manner will have an encouraging impact on future neural field work.  
As a result, on the large indoor scene ({\bf Lobby}), our method achieves the state-of-the-art projection quality of the neural surface field, even with 83.3\% reduced number of samples. 
% Additionally, we tested our approach on NeRF \cite{nerf} as well, which has less accurate geometry. 
% It turned out that, with our recovery algorithm, our method can render $2.8 \times$ faster than \cite{nerf}, with negligible performance differences. 
% Our recovery algorithm realizes this by rendering few rays again with the full scene boundary, only if the sum of the weight along that ray is lower than the threshold. 

\subsection{Limitations and Future Work}
Even though we have proposed a rigorous policy to handle the failures to find correct surfaces, falling back to the entire ray range may be suboptimal. 
A good way to potentially mitigate this problem would be to use multi-resolution approaches, such as octree-based representations \cite{adaptivepleoctree,voxel4}, so that the maximum resolution of the discrete TSDF volume can increase without much memory consumption. 
Furthermore, we notice that the geometry of the model tends to converge in the middle of training. 
Therefore, we envision our future research direction as extending our method into the main training process with a similar insight, so that we are not limited to boosting the inference efficiency, but can further accelerate the training process.

{
    \small
    \bibliographystyle{ieeenat_fullname}
    \bibliography{main}
}

% \appendix
% \renewcommand{\thesection}{\Alph{section}}
% \setcounter{section}{0}
% \begin{appendices}
\clearpage
\setcounter{page}{1}
\maketitlesupplementary
\label{sec:suppl}
In the \textbf{main paper}, we presented a model-agnostic approach for efficient rendering of neural surface fields. Unlike previous fast rendering of neural surface fields~\cite{neus2} that handled objects only, our method proved to be able to generalized into scenes. In this \textbf{supplementary document}, we additionally show the need of TSDF-Sampling through the implementation details (Sec.~\ref{supp: implementation_detail}), additional ablations (Sec.~\ref{supp: additional_ablation}), and additional geometric results (Sec.~\ref{supp: additional_geometric}).
% show encouraging effects on

\section{Implementation Detail}
\label{supp: implementation_detail}
\subsection{TSDF Integration}
Algorithm \ref{alg:integration} illustrates the integration process described in Section \ref{sec:integration}. TSDF grid $\mathcal{V}$ and and TSDF weight grid $\mathcal{W}$ are initialized by -1, which means unseen voxels, and 0, respectively.  In the Algorithm \ref{alg:integration}, we incrementally update the voxel $\mathcal{V}(X)$, following the rule from \cite{volrange,voxblox}. For updating the $\mathcal{W}(X)$, we were inspired by the function introduced in  \cite{realtime,volrange}.

\SetKwInput{kwInput}{Input}
\SetKwInput{kwRequire}{Require}
\SetKwInput{kwOutput}{Output}
\SetKwInput{kwReturn}{Return}
\begin{algorithm}[h]
\caption{TSDF integration}
\DontPrintSemicolon
\kwInput{\\
$\mathcal{P}$: set of rays\\
$\mathcal{B}$: NDC \\
$\bold{o}_i$: camera origin of i-th frame \\
$\bold{v}_{ij}$: j-th ray direction of i-th frame \\ 
$\hat{D}_{ij}$: j-th ray depth of i-th frame \\   

\kwRequire{\\
% $t_n$: near bound of $t$ to sample\\
% $t_f$: original far bound of $t$\\
$D_{T}$: truncated distance\\
}
\kwOutput{\\
updated TSDF volume $\mathcal{V}$\\ 
updated weight volume  $\mathcal{W}$
}\;

\For{$(\bold{o}_i, \bold{v}_{ij}, \hat{D}_{ij}) \in \mathcal{P}$}{
    $\bold{p}^* \gets \bold{o}_i+\hat{D}_{ij}\bold{v}_{ij}$\\
    $t \gets 0$\;
    $\bold{p} \gets \bold{o}_i+t\bold{v}_{ij}$\;
    $X \gets \textsc{Voxel}(\bold{p}$)\\
    \While{$\bold{p} \in \mathcal{B}$}{
        $\bold{c}_{X} \gets \textsc{VoxelCenter}(X)$ \\
        $s_{X} \gets \textsc{clamp}(\bold{v}_{ij} \cdot (\bold{p}^*-\bold{c}_X),-D_T,D_T) $ \\
        \eIf{$s_{X} > - D_T$}{
            \\$w_{X} \gets \textsc{Weight}(s_{X})$ \\
            $\mathcal{V}(X) \gets \dfrac{\mathcal{W}(X)\mathcal{V}(X)+w_{X}s_{X}}{\mathcal{W}(X)+w_{X}}$ \\
            $\mathcal{W}(X) \gets \mathcal{W}(X)+w_{X}$\\
            % $X,\bold{p}_{k+1} \gets \textsc{NextVoxel}(\bold{p}_k,\bold{v}_{ij})$
            $X, t \gets \textsc{NextVoxel}(t,\bold{v}_{ij})$\;
            $\bold{p} \gets \bold{o}_i+t\bold{v}_{ij}$
        }{
        \textbf{break}
        }
    }   
}
}
\kwReturn{
$\mathcal{V}$, $\mathcal{W}$
}
\label{alg:integration}
\end{algorithm}

\begin{figure}[!tp]
    \renewcommand{\tabcolsep}{0.5pt}
    \centering \footnotesize
    \begin{tabular}{c}
        \includegraphics[width=0.42\textwidth, height=0.18\textwidth]{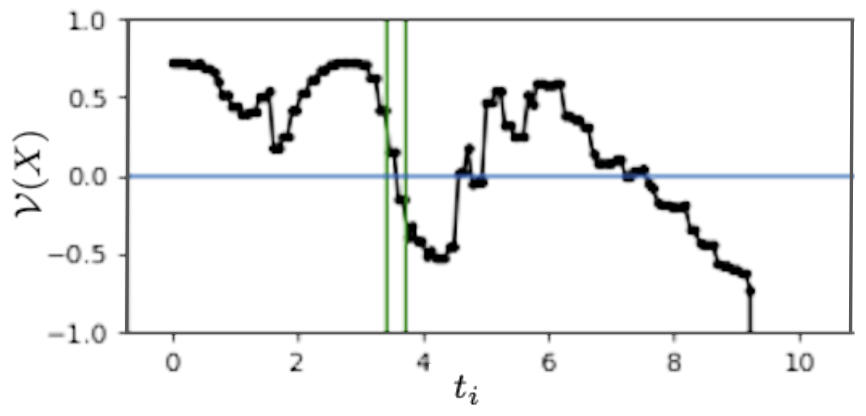}
    \end{tabular}
    \caption{The TSDF values stored in an array of voxels, that a ray passes through. The green lines represent $t_n$ and $t_f$ of our method.}
    \label{fig: suppl_tsdfray}
\end{figure}

In the Algorithm \ref{alg:spacecarving}, the TSDF voxels that $\bold{p}_k$ visit give SDF values $\mathcal{V}(X)$. Fig. \ref{fig: suppl_tsdfray} displays $\mathcal{V}(X)$ along a ray. Note that Algorithm \ref{alg:spacecarving} stops visiting the voxels when $t_f$ is found, but Fig. \ref{fig: suppl_tsdfray} shows $\mathcal{V}(X)$ until the ray reaches \textit{unseen} area, i.e., -1, just for visualization purposes. Since the ray goes through empty spaces, then meets the edge of the vending machine, eventually goes out of the vending machine, and lastly hit the wall behind it, the plot of the $\mathcal{V}(X)$ is drawn as Fig. \ref{fig: suppl_tsdfray}. The green lines represent the $t_n$ and $t_f$. These effectively shows that our algorithm set $t_n$ at the point when the ray starts to be near a surface and $t_f$ at the point when the ray is discovered to be sufficiently inside an object.

\subsection{Training Details for Lobby and Garage}
\begin{figure}[!tp]
    % \centering
    \begin{tabular}{cc}
    \includegraphics[width=0.2\textwidth]{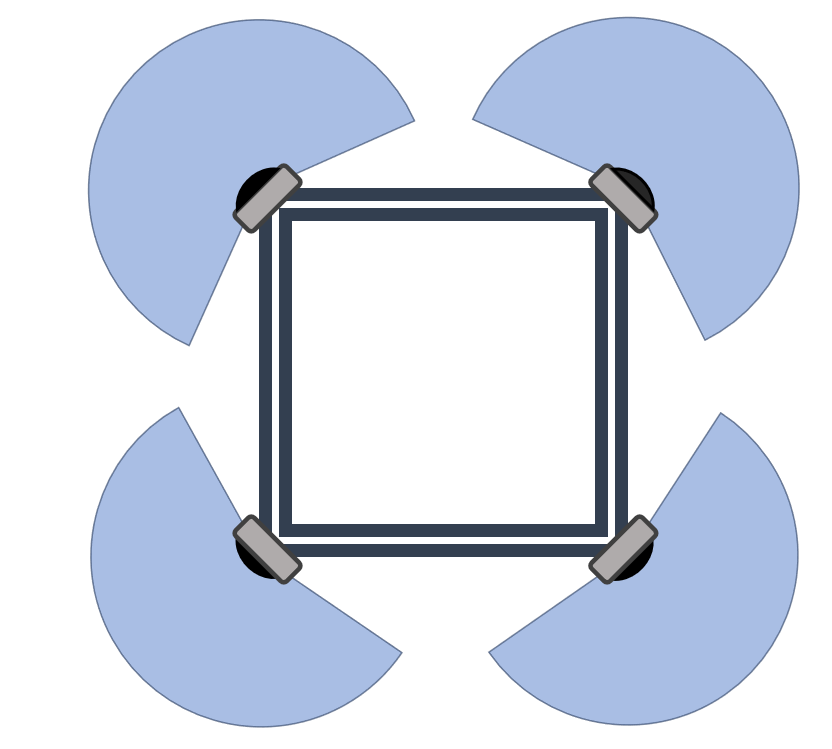} & 
    \includegraphics[width=0.17\textwidth]{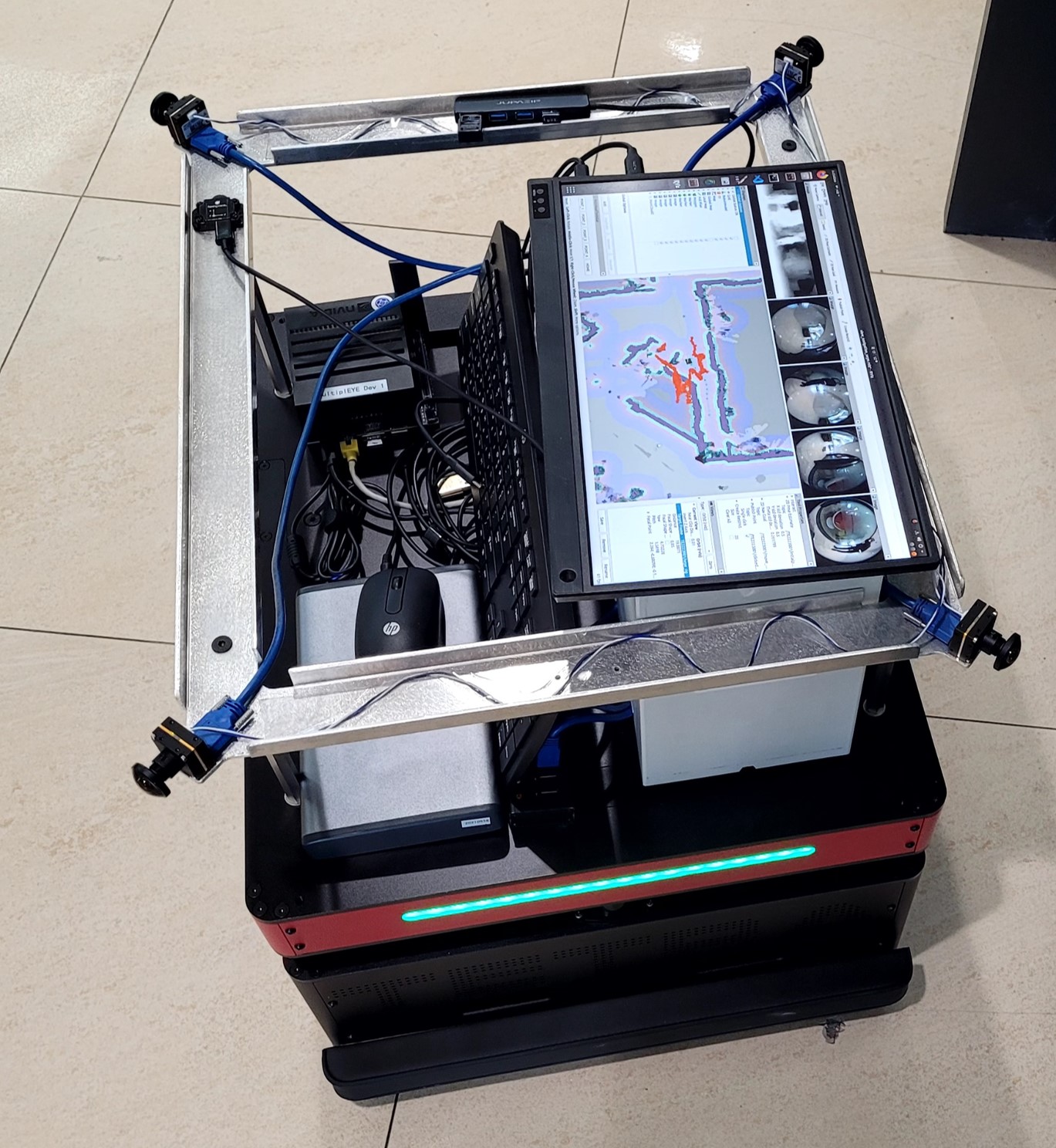} \\
     (a) & (b) \\
    \end{tabular}
    \caption{(a): Omnidirectional multi-camera rig system \cite{omnimvs} (b): $\bold{Lobby}$ was recorded by the mobile robot equipped with the system in (a).}
    \label{fig:camera_setup}
\end{figure}

Fig. \ref{fig:camera_setup} shows the camera rig system for the Lobby and Garage dataset in (a).  The mobile robot shown in (b) captured the Lobby dataset. The data capture system and the guide depth for the training was obtained by using~\cite{omnimvs}, which can be suitable for collecting data in a large scene.
% Since~\cite{omnimvs} returns the depth map in Since the output of \cite{omnimvs} is an omnidirectional depth map, we warped it to train camera view. \\

% \subsubsection{Training Loss}
We used the same loss functions from the official implementation of our baselines to optimize the network, except for the depth loss. This is because the depth of the datasets we used are in meters, so we did not need to use the monocular scale and shift presumption when computing the depth loss. 
We mapped the depth to disparity as~\cite{sweepnet} and guided the network with the following loss function, inspired by~\cite{dsnerf}, 
% Since the error over far distance is usually large, to avoid overfitting in that area, we design the loss function by converting ground-truth depth $D^*$ and the depth computed by Eq. 
% \ref{eq:compute_depth}, $D$, in the disparity space\cite{sweepnet} as:   
\begin{equation}
\mathcal{L}_{depth} = \sum_{(\bold{o},\bold{v}) \in \mathcal{P}} \Vert \hat{d}_\mathcal{T}(\bold{o},\bold{v}) - d^*(\bold{o},\bold{v})\Vert^2,
\label{eq:depthloss}
\end{equation}
where $(\bold{o},\bold{v})$ implies a ray in the train set $\mathcal{P}$. $d(\cdot) = 1/D(\cdot)$ denotes disparity, and the $D(\cdot)$ defines depth. 

% \begin{equation}
% \mathcal{L} = \mathcal{L}_{photo} + \lambda_{eik}\mathcal{L}_{eik} + \lambda_{depth}\mathcal{L}_{depth} + \lambda_{normal}\mathcal{L}_{normal} + \lambda_{smooth}\mathcal{L}_{smooth} 
% \label{eq:total_loss}
% \end{equation}

\section{Additional Ablation Results}
\label{supp: additional_ablation}
\subsection{Adaptive Sampling}
\begin{figure*}
    \renewcommand{\tabcolsep}{1.3pt}
    \centering
    \begin{tabular}{ccccc}
    \includegraphics[width=0.18\textwidth]{Tetra/zoom_in/zoom_in_Input_image.png} &
        \includegraphics[width=0.18\textwidth]{Tetra/zoom_in_depth/zoom_in_hs_96.png} &
        \includegraphics[width=0.18\textwidth]{Tetra/zoom_in_depth/zoom_in_hs_14.png} &
        \includegraphics[width=0.18\textwidth]{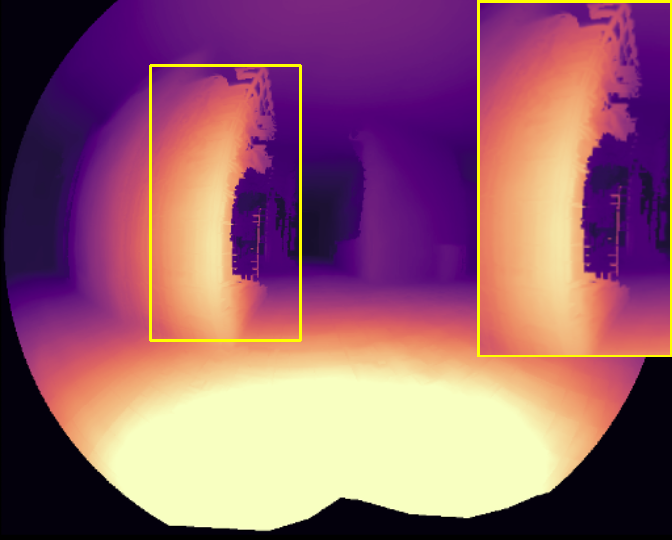} &
        \includegraphics[width=0.18\textwidth]{Tetra/zoom_in_depth/zoom_in_TSDF_full_14.png} \\
        \includegraphics[width=0.18\textwidth]{Garage/zoom_in_color/zoom_in_gt_image.png} &
        \includegraphics[width=0.18\textwidth]{Garage/zoom_in_depth/zoom_in_resampling_96_depth.png} &
        \includegraphics[width=0.18\textwidth]{Garage/zoom_in_depth/zoom_in_resampling_14_depth.png} &
        \includegraphics[width=0.18\textwidth]{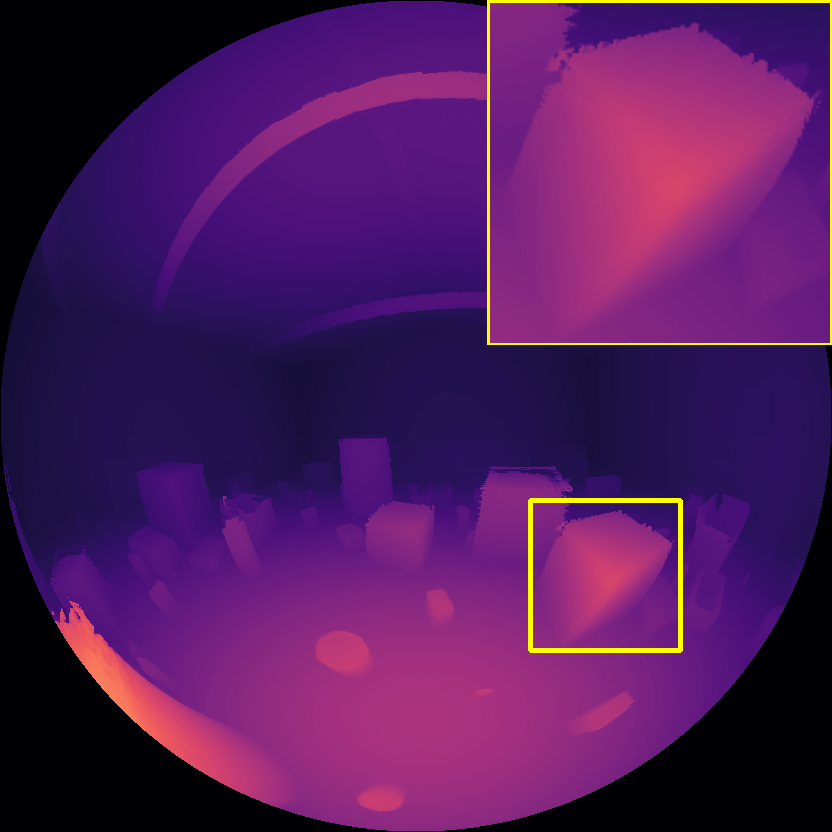} &
        \includegraphics[width=0.18\textwidth]{Garage/zoom_in_depth/zoom_in_ours_14_depth.png} \\
        \includegraphics[width=0.18\textwidth]{Replica/unnorm_fidx70_rect/zoom_in_000070_rgb.png} &
        \includegraphics[width=0.18\textwidth]{Replica/unnorm_fidx70_rect/zoom_in_base_sampler_fidx_70_inv_depth.png} &
        \includegraphics[width=0.18\textwidth]{Replica/unnorm_fidx70_rect/zoom_in_base_sampler_16_32_fidx_70_inv_depth.png} &
        \includegraphics[width=0.18\textwidth]{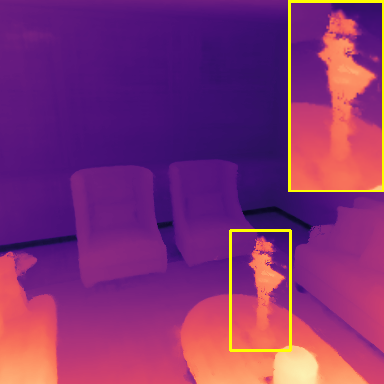} &
        \includegraphics[width=0.18\textwidth]{Replica/unnorm_fidx70_rect/zoom_in_tsdf_sampler_adaptive_48_fidx_70_inv_depth.png} \\
        (a)  & (b)  & (c)  & (d)  & (e) \\
    \end{tabular}
    \caption{Ablation on our adaptive algorithm. (a): Ground-Truths (b): Pre-trained model with a large number of samples. (c): Pre-trained model with a limited number of samples. (d): Ours (naive) with a limited number of samples. (e): Ours (with the adaptive algorithm) with a limited number of samples.}
    \label{fig:suppl_qualitative}
\end{figure*}

In Fig. \ref{fig:suppl_qualitative}, we present the qualitative results on our ablation study between our naive method and our full method. Without our TSDF-Sampling constraint, (c) incurs a notable loss of object surfaces. (d) mitigates this error by using the TSDF-Sampling, but the edges of objects still shows losses. In (e), on the other hand, our approach shows compelling effect, and the objects are correctly rendered, with around 7 times less number of samples than (b). 
\subsection{Analysis on Adaptive Sampling}
\begin{figure*}[!tp]
    \renewcommand{\tabcolsep}{0.5pt}
    \centering \footnotesize
    % \begin{tabular}{ccc}
    % \includegraphics[width=0.35\textwidth, height=0.3\textwidth]{Raydist/Hard/Baseline_16.png} &
    %  \includegraphics[width=0.32\textwidth,height=0.3\textwidth]{Raydist/Hard/Ours_16_wo_adaptive.png} &
    %  \includegraphics[width=0.34\textwidth,height=0.3\textwidth]{Raydist/Hard/Ours_32.png} \\    
    % \includegraphics[width=0.34\textwidth, height=0.3\textwidth]{Raydist/Easy/Baseline_16.png} &
    %  \includegraphics[width=0.34\textwidth,height=0.3\textwidth]{Raydist/Easy/Ours_16_woadaptive.png} &
    %  \includegraphics[width=0.34\textwidth,height=0.3\textwidth]{Raydist/Easy/Ours_11.png} \\
    %  (a) \cite{monosdf} with 16 samples in average &(b) Ours (naive) with 16 samples in average &(c) Ours (full) with 16 samples in average. 
    % \end{tabular}
    \begin{tabular}{ccc}
    
     \includegraphics[height=0.1458\textwidth]{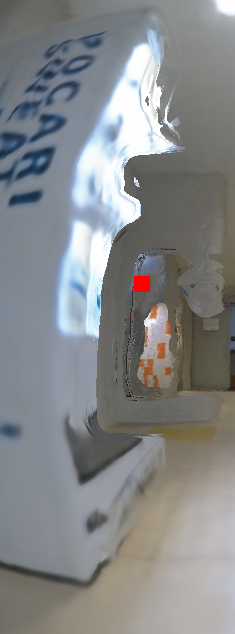} &
    \includegraphics[height=0.1458\textwidth]{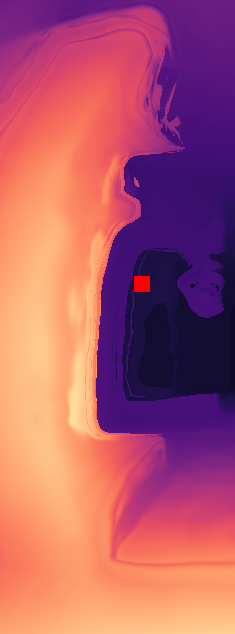} &
    \includegraphics[width=0.7\textwidth,height=0.1458\textwidth]{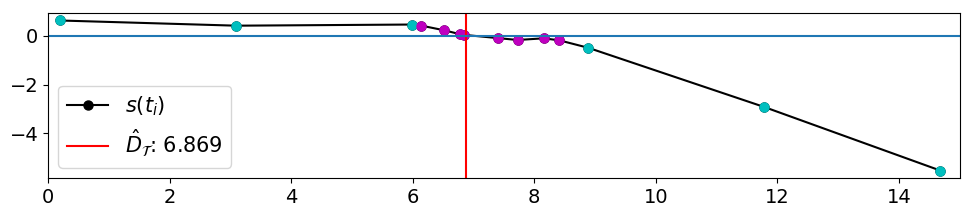} \\  % 0.1458 0.1250
    &&(a) Hierarhical Sampling on ~\cite{monosdf}, 14 samples\\
    \includegraphics[height=0.1458\textwidth]{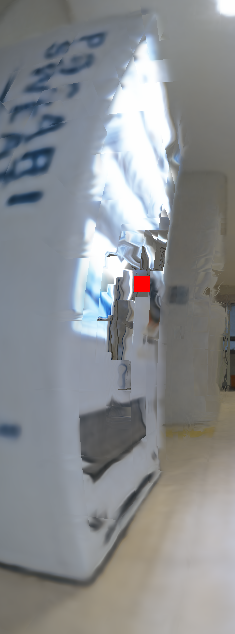} &
    \includegraphics[height=0.1458\textwidth]{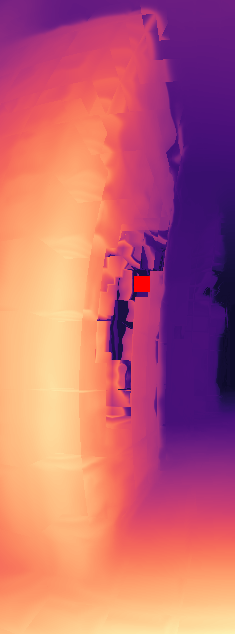} &
    \includegraphics[width=0.7\textwidth,height=0.1458\textwidth]{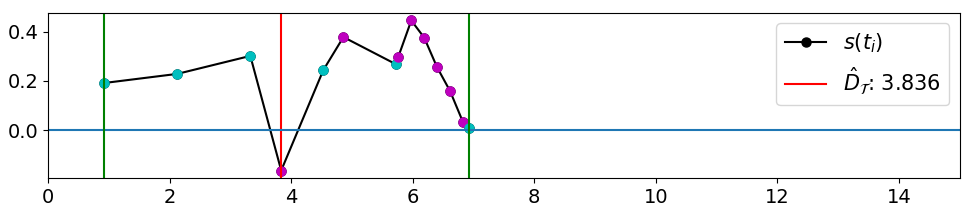} \\
    &&(b) TSDF-Sampling (naive) on ~\cite{monosdf}, 14 samples\\
     \includegraphics[height=0.1458\textwidth]{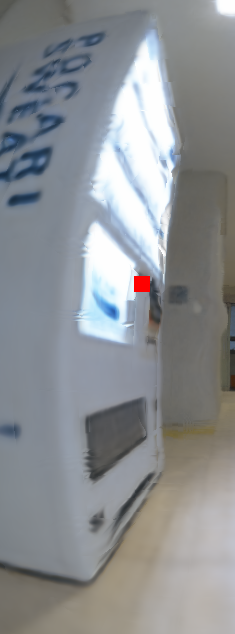} & 
     \includegraphics[height=0.1458\textwidth]{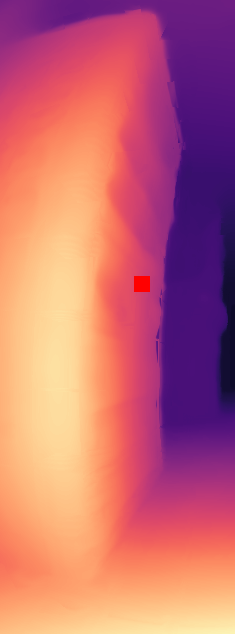}&
    \includegraphics[width=0.7\textwidth,height=0.1458\textwidth]{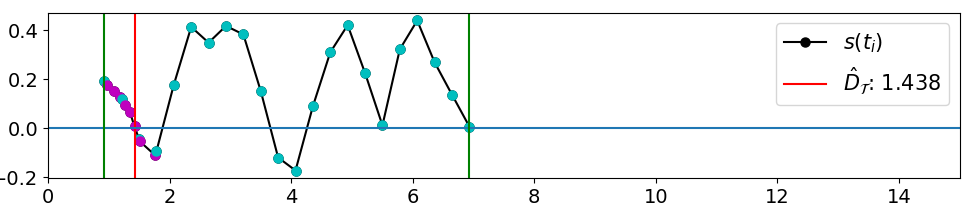} \\
     &&(c) TSDF-Sampling (full) on ~\cite{monosdf}, 31 samples for this ray (14 samples in average across the scene)\\\\

     \includegraphics[width=0.05404\textwidth,height=0.1458\textwidth]{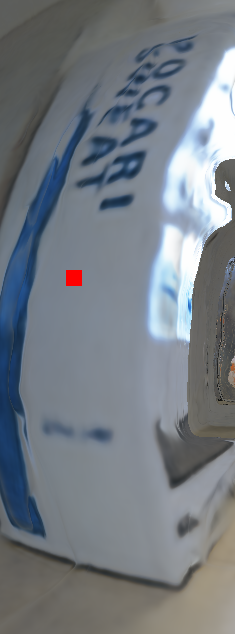} &
    \includegraphics[width=0.05404\textwidth,height=0.1458\textwidth]{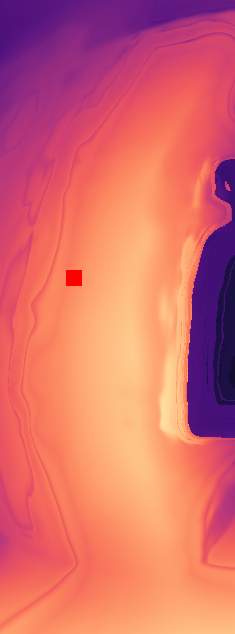} &
    \includegraphics[width=0.728\textwidth,height=0.1458\textwidth]{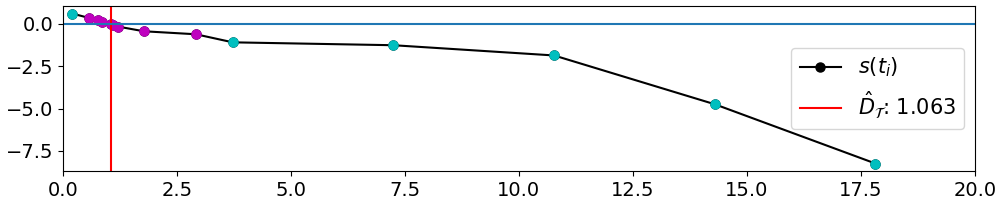} 
    \\ 
    &&(d)Hierarhical Sampling on ~\cite{monosdf}, 14 samples\\
    \includegraphics[width=0.05404\textwidth,height=0.1458\textwidth]{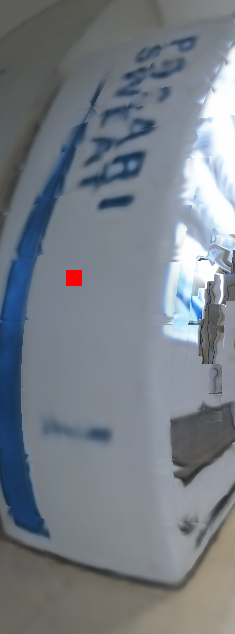} &
    \includegraphics[width=0.05404\textwidth,height=0.1458\textwidth]{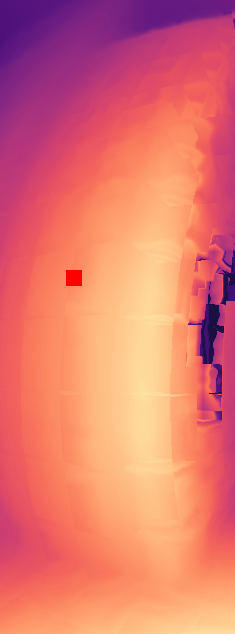} &
    \includegraphics[width=0.728\textwidth,height=0.1458\textwidth]{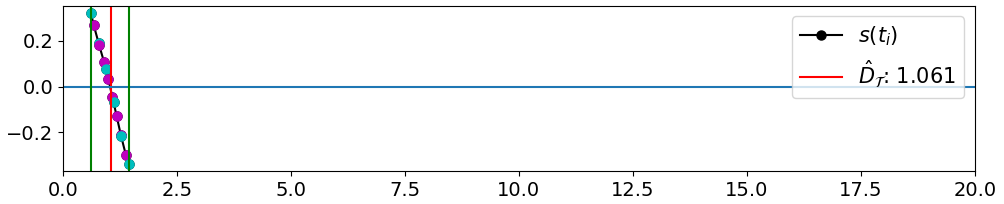} 
    \\
    &&(e) TSDF-Sampling (naive) on ~\cite{monosdf}, 14 samples\\
     \includegraphics[width=0.05404\textwidth,height=0.1458\textwidth]{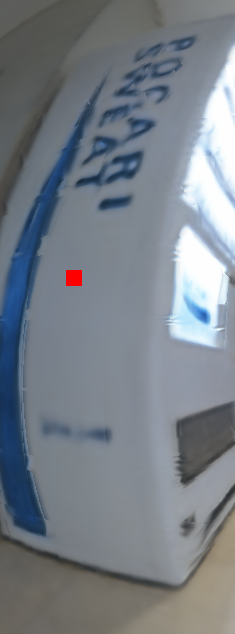} & 
     \includegraphics[width=0.05404\textwidth,height=0.1458\textwidth]{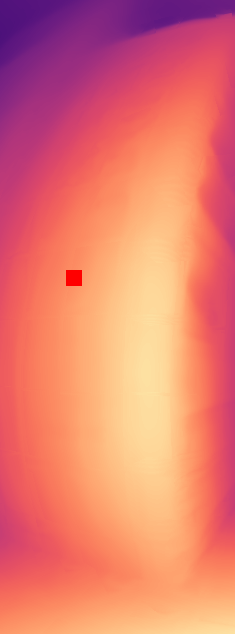}&
    \includegraphics[width=0.728\textwidth,height=0.1458\textwidth]{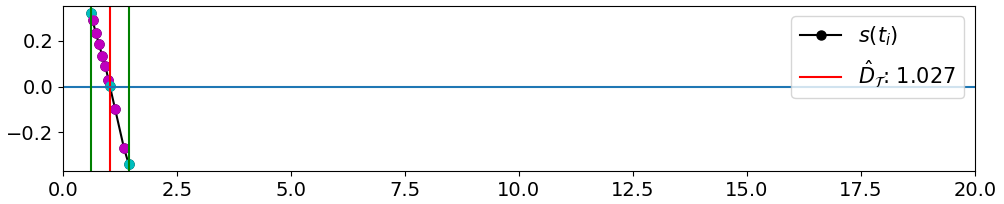} 
    \\
     &&(f) TSDF-Sampling (full) on ~\cite{monosdf}, 11 samples for this ray (14 samples in average across the scene)\\
     
    \end{tabular}
    \caption{Justification of the need for the TSDF-Sampling on the neural surface field models. (a)-(c) shows the challenging case when a ray hits an edge of an object, where the true surface fails to be detected by the reduced number of samples in (a) and (b). In (c), our adaptive TSDF-Sampling automatically sets a narrower band and more samples for this particular ray, achieving successful rendering of the vending machine. (d)-(f) shows a ray hitting a plain surface, where (e) and (f) are more focused on the surface than (d), and (f) automatically uses less samples for this less challenging ray.  
    % The first row analyzes an edge case, while the second row shows a planar case. The samples in the coarse stage are drawn in magenta markers and the samples in the fine stage are drawn in cyan markers. Best viewed zoomed in.
    }
    \label{fig:suppl_raydist}
\end{figure*}

In addition to the Fig. \ref{fig:raydist} in Sec. \ref{sec:exp}, we present further analysis on the SDF distribution on rays, with respect to $t_i$. While Fig. \ref{fig:raydist} shows a ray that penetrates the vending machine, we show the cases of casting the ray to a plane and to an edge in Fig. \ref{fig:suppl_raydist}. The cyan markers represent the coarse samples, and the magenta markers indicate the fine samples. Fine samples are upsampled from the PDF of the coarse samples by using the inverse CDF~\cite{nerf,volsdf}. $s$ denotes the SDF values, and the rendered depths $\hat{D}_r$ on rays $r$ are in meters. The green lines shows $t_n$ and $t_f$ of the TSDF-Sampling approach.

In (a), because of the large scene size, the sampling range is more than 14m, so the gaps between coarse samples are more than 2m. This makes the Hierarchical Sampling prone to skip the true surfaces. The result on the left of the plot demonstrates the consequence. On the other hand, the TSDF-Sampling in (b) restrict the sampling bound within the green lines. As a result, the 6 coarse samples are more meaningfully located inside the bound, so they are able to be denser. Hence, the rendering result shows less lossy image and depth map. In (c), the TSDF-Sampling approach boost the performance even further with our adaptive sampling module added. 5 meters of the sampling bound is made by the complex geometry at the edge of the vending machine, and 5 meters is larger than the average sampling bound. Therefore, our adaptive sampling approach automatically increases the number of samples in this ray, so that it can handle this challenging ray with more neural network queries. This approach results in robust quality of the challenging edge cases, shown in the left of the plot. 

In (d)-(f), the case of plain geometry is shown. The main difference from (a)-(c) is that this ray does not have multi-planes. As a result, even though the first and the second coarse sample in (d) has more than 2.5 meters of gap, the zero-crossing between the first and the second sample was the only zero-crossing and also was the only true surface. Therefore, the concentrated fine samples between those two coarse samples were a correct thing, resulting in finding the true surface. Our method shown in (e) makes it even safer by avoiding the sampling in random empty space. The coarse samples in (e) are focused between the green line. However, we can observe that such narrow bound does not necessarily need 14 samples. In (f), our adaptive approach automatically reduces the number of coarse samples from 6 to 3, because of the narrower bound than the average bound length. Our approach is able to save even more time on this less challenging ray, as shown in (e). 

\subsection{Recovery Algorithm}
We evaluate the performance of the Recovery algorithm, as detailed in Section \ref{subsection:recovery}, by adjusting the threshold. Tab. \ref{tab:suppl_recovery} illustrates that, with an increasing threshold, the photometric performance (PSNR and SSIM) either remains steady or slightly surpasses the performance of the original sampling method. Nevertheless, the rise in rendering time and total samples become more significant, outweighing the performance gains, especially when the threshold approaches 1. This happens because as the threshold increases, it re-renders pixels that have already been fully rendered. Conversely, when the threshold is too small, it leads to a significant drop in performance, thus, we set it to 0.95.

\section{Additional Geometric Results}
\label{supp: additional_geometric}

\subsection{Normal Estimation}
\begin{figure*}[!tp]
    \renewcommand{\tabcolsep}{1.1pt}
    \centering \footnotesize
    \begin{tabular}{cccc}
    \includegraphics[width=0.2\textwidth]{Tetra/zoom_in/zoom_in_Input_image.png} &
    \includegraphics[width=0.2\textwidth]
    {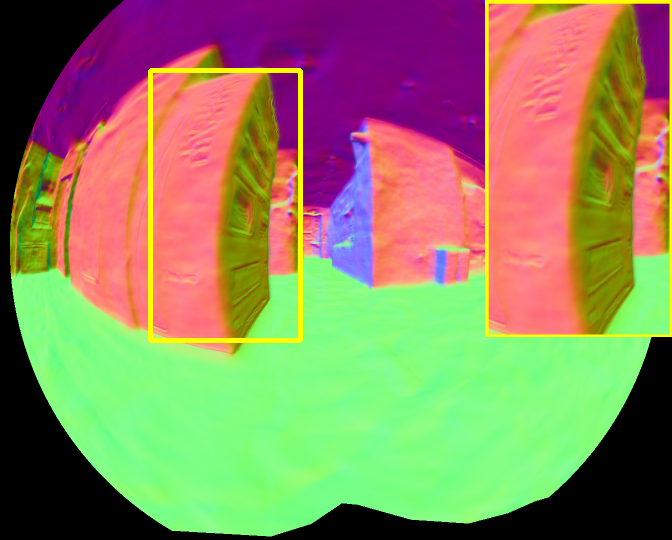} &
    \includegraphics[width=0.2\textwidth]
    {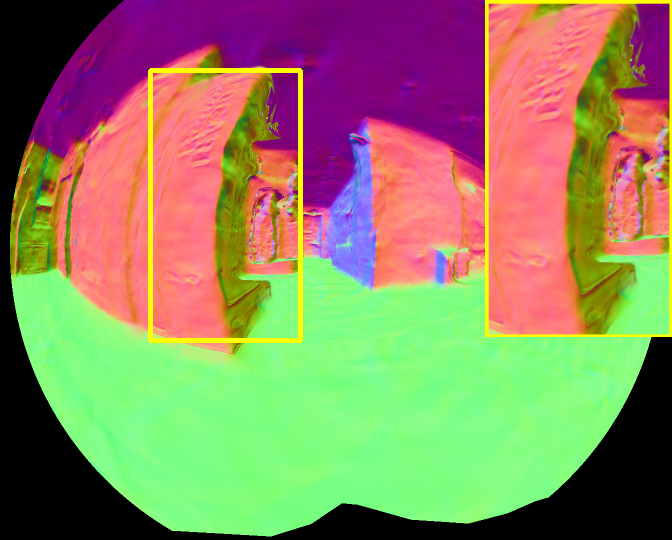} &
    \includegraphics[width=0.2\textwidth]
    {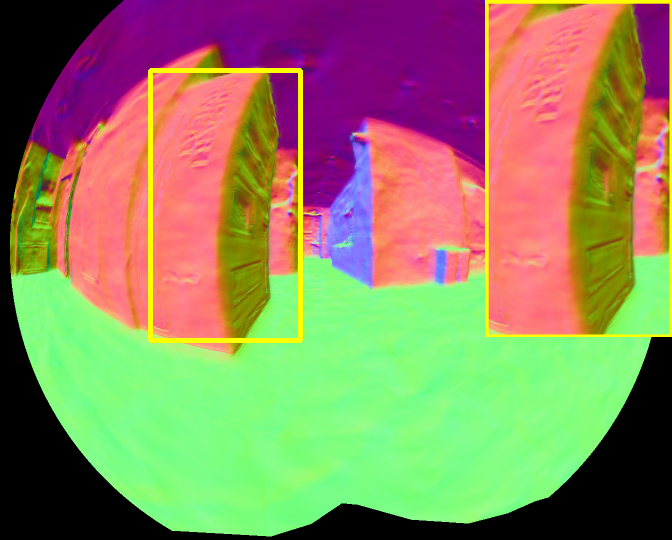} \\
    \includegraphics[width=0.2\textwidth]{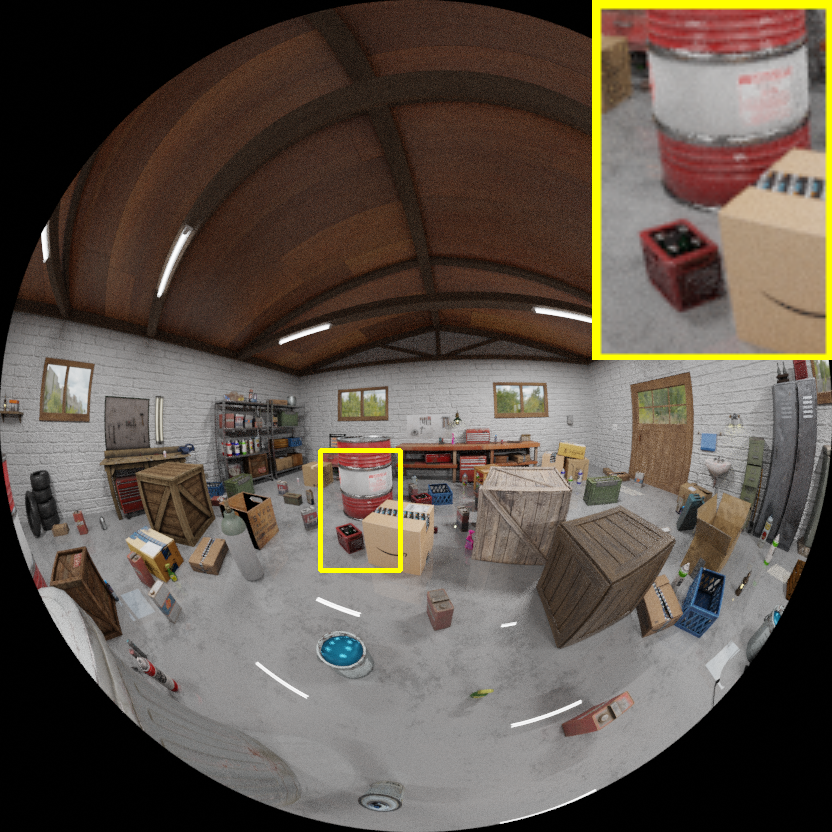} &
    \includegraphics[width=0.2\textwidth]{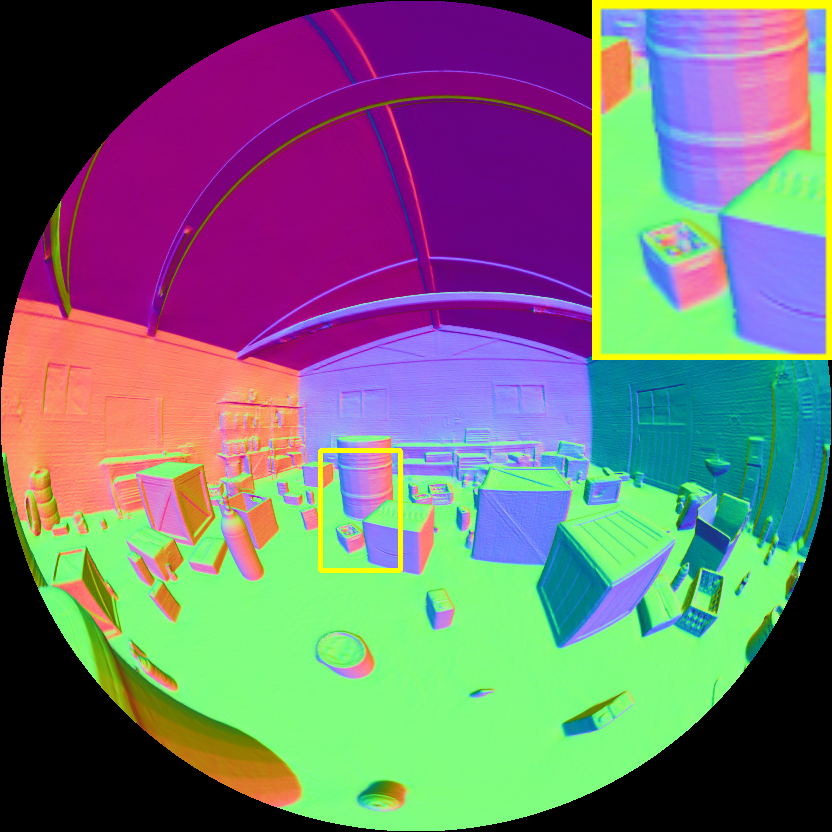}&
    \includegraphics[width=0.2\textwidth]{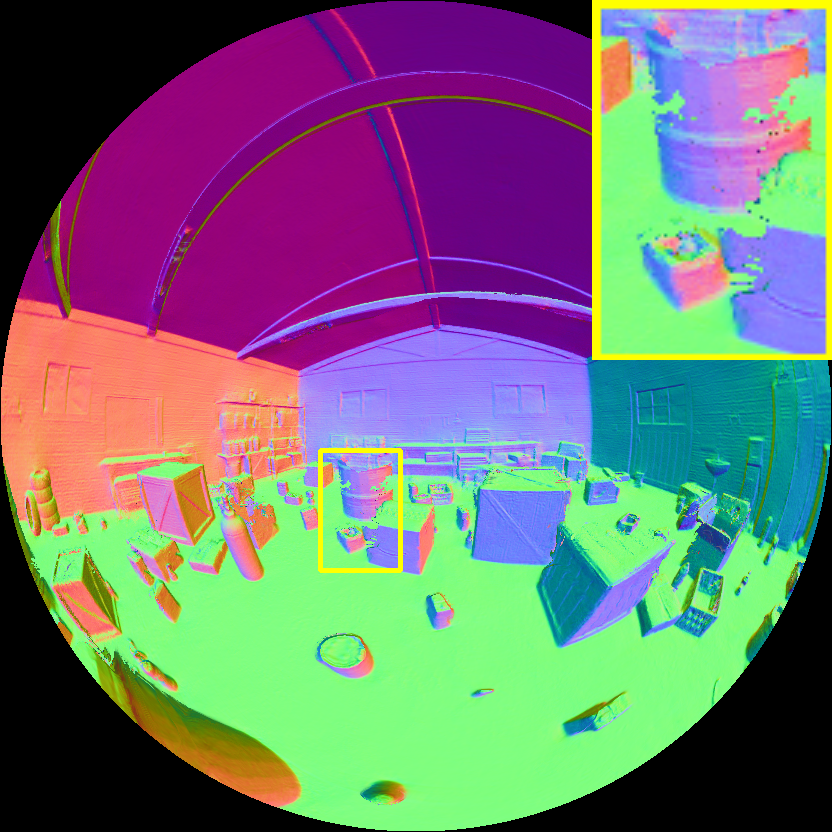}&
    \includegraphics[width=0.2\textwidth]{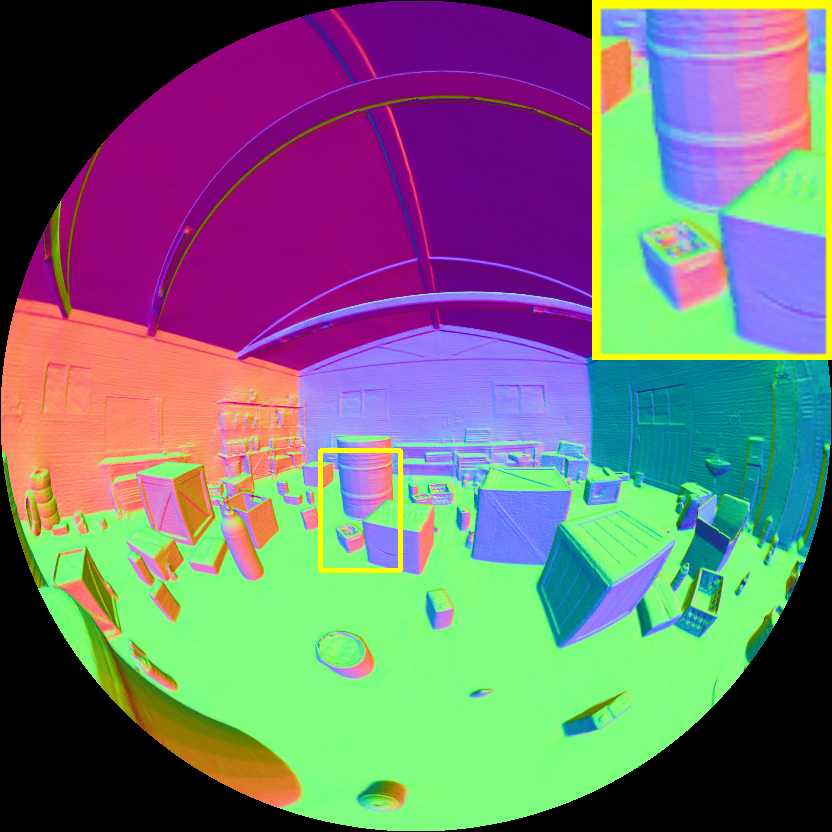}\\
    (a) & (b) & (c) & (d)\\
    \end{tabular}
    \caption{Normal estimation from the neural surface field model. (b) and (c) used the Hierarchical Sampling on MonoSDF~\cite{monosdf}, with 96 and 14 samples, respectively. (d) used our TSDF-Sampling approach on the~\cite{monosdf} model with only 14 samples. This simple TSDF-Sampling approach saves the rendering time by 6.86 times. Since the TSDF-Sampling is model-agnostic, this gain can be easily generalized to any existing neural surface field networks as long as they use volume rendering. We displays the input image in (a).}
    \label{fig:suppl_normal}
\end{figure*}

Our method achieves high fidelity on both depth estimation and normal estimation. In Fig. \ref{fig:suppl_normal}, we present the estimated normal map. In the volume rendering process, we calculate the rendered normal map $\hat{N}$ as follows: 
\begin{equation}
\hat{N}_\tau(\bold{o}, \bold{v}) = \frac{1}{\sum_{t_i \in \mathcal{T}} w(t_i)} \sum_{t_i \in \mathcal{T}} w(t_i) \bold{n}(\bold{p}(t_i)),
\label{eq: supp_normal}
\end{equation}
where the surface normal $\bold{n}(\bold{p}) = \nabla_\bold{p} f_\theta(\bold{p}) / ||\nabla_\bold{p} f_\theta(\bold{p})||_2 $ at a position $\bold{p}$ is the normalized value of the analytical gradient of the neural surface field network $f_\theta$. While the neural surface field baseline produces reasonable results, the results can be reproduced with less samples, only when advanced sampling strategy like ours is applied.

% \subsection{Comparison with MVS}
% Since the Lobby dataset does not provide the model with a sensor depth, we compare the geometric performance of our method with regards to the classical deep Multi-view Stereo (MVS) algorithm. 
% In Tab. \ref{tab:suppl_lobby} shows the depth and normal map comparison with a MVS \cite{omnimvs} results. Although the MVS result also have its error, the trend of our method and \cite{monosdf} gets distinct when the number of samples is reduced. 
% Notably, although our TSDF-Sampling is not related to the MVS approach, our method on~\cite{monosdf} outperforms in 14 cm of depth, compared to the Hierarchical Sampling on the same backbone model in 6+6 samples. Normal shows the same trend with depth, ours outperforming more than 2 degree when the reduced number of samples were used. 
% Our method outperforms \cite{monosdf} in a significant amount in geometry.
% With 96 samplings per ray, the performance between \cite{monosdf} and ours are not significant. However, when the limited number of samples are used to boost the model, ours outperforms \cite{monosdf} in a significant amount, especially in geometry

\section{Detailed Experimental Results}
Tab. \ref{tab:suppl_garage} reports quantitative comparisons in more detailed variations of the number of samples. This supports the drastic improvements of PSNR, depth MAE, and normal angle error of the TSDF-Sampling upon the existing methods, shown in Fig.~\ref{fig:samples_vs_performance}. 
The number of samples for our method (full) has been rounded off for simplicity. 
The minor differences of the rendering time within the same number of samples is due to the use of CUDA in ours, but the tendency of rendering time to be inverse proportional to the number of samples remains valid.
The depth error for our method (full) even becomes slightly better when given less average number of samples. 
This signifies that our method has a potential to even beat the exhaustive samplings. This might be because our reduced $\Delta t$ makes the samples be very dense, so the samples eventually become denser than the finite training data can afford, when given too many samples. 
% We envision that if the neural surface field model has been trained with even denser rays, 
The gap between coarse samples, $\Delta t\text{[cm]}$, effectively shows the reason why our TSDF-Sampling outperforms significantly with the limited number of samples. The narrow band the TSDF-Sampling sets, based on the TSDF prior, prevents the samples from being overly sparse.  

\begin{table*}[h]
    \centering
    \begin{tabular}{c|ccrccc}
        \toprule
        $\tau$ & Total Samples &$\sum w < \tau$& Time [s] $\downarrow$& 
        PSNR [dB] $\uparrow$ & SSIM $\uparrow$\\
        \midrule
        0.99 & 120  & 37.76  & 3.30 & 34.21 & 0.995 \\
        0.98   & 106  & 30.39 & 2.81 & 34.14 & 0.995 \\
        0.95  & 86  & 20.15 & 2.42 & 33.65 & 0.995\\
        0.85     & 65  & 9.24 & 1.87 & 31.39 & 0.992 \\
        0.75    & 58  & 5.13 & 1.61 & 29.84 & 0.989\\
        \bottomrule
    \end{tabular}
\caption{Performance comparison of threshold $\tau$ for recovery algorithms based on average sample number of 48 without recovery applied. Total Samples shows the average number of samples, including the number of samples taken during the recovery process.}
    \label{tab:suppl_recovery}
\end{table*}

 % With this simple plug-and-play approach, we can save

% \begin{figure*}[!tp]
%     \renewcommand{\tabcolsep}{1.3pt}
%     \centering \footnotesize
%     \begin{tabular}{cccc}
%         \includegraphics[width=0.236\textwidth,height=0.197\textwidth]{Tetra/view1/Input_image.png} &
%         \includegraphics[width=0.235\textwidth,height=0.198\textwidth]{Tetra/view1/hs_14.png} &
%         \includegraphics[width=0.24\textwidth]{Tetra/view1/TSDF_naive_14.png} &
%         \includegraphics[width=0.24\textwidth]{Tetra/view1/TSDF_full_14.png} \\
%         \includegraphics[width=0.18\textwidth]{no_image.png} &
%         \includegraphics[width=0.235\textwidth]{Tetra/view1_depth/hs_14.png} &
%         \includegraphics[width=0.24\textwidth]{Tetra/view1_depth/TSDF_naive_14.png} &
%         \includegraphics[width=0.24\textwidth]{Tetra/view1_depth/TSDF_full_14.png} \\
        
%         GT & MonoSDF 16 & Ours 16 w/o adaptive & Ours 16 \\
%     \end{tabular}
%     \caption{Ablation results on our adaptive component}
%     \label{fig:ablation_tetra}
% \end{figure*}

\begin{table*}[htbp]
    \centering
    \renewcommand{\arraystretch}{1.1}
    \scalebox{0.9}{
    \begin{tabular}{c|ccrrrrrrr}
        \toprule
          Approaches & Sampling &Samples &$\Delta t$ [cm]& Time [s]$\downarrow$  & PSNR [dB] $\uparrow$  & SSIM $\uparrow$ & D [cm]$\downarrow$  & N [deg]$\downarrow$  \\
        \midrule
            Mip-NeRF \cite{mipnerf} &HS& 64+32 &17.20& 28.00 & 27.35 & 0.984 & 8.42 & 32.2\\
            % Instant-NGP\cite{ngp}  &RM& 31 &-& 3.30   & 27.84 & 0.977 & 8.64 & 27.5\\
            MonoSDF \cite{monosdf}  &EB& 64+32 &17.20& 54.76 & 26.90 & 0.979 & 9.36 & 6.73\\
            \hline
            MonoSDF \cite{monosdf}        &HS& 64+32&17.20 & 19.62 & 28.05& \textcolor{silver}{0.980}& \textbf{\textcolor{gold}{8.72}}& 6.84\\
            Ours (naive)             &TSDF& 64+32&0.97 & 18.06 & \textbf{\textcolor{gold}{28.06}}& \textbf{\textcolor{gold}{0.981}}& \textcolor{silver}{8.89}& \textcolor{silver}{6.77}\\
            Ours (full)              &TSDF&64+32 &0.97&18.27 & \textcolor{silver}{28.06}& \textcolor{silver}{0.980} & 8.90 &\textbf{\textcolor{gold}{ 6.76}} \\
            \hline
            MonoSDF \cite{monosdf}        &HS& 16+32 &68.95& 10.78&27.52&0.978&\textbf{\textcolor{gold}{8.58}}&7.14\\
            Ours (naive)             &TSDF& 16+32 &3.48& 11.11&\textcolor{silver}{28.05}&\textcolor{silver}{0.980}&8.89&\textcolor{silver}{6.79}\\
            Ours (full)              &TSDF& 16+32 &3.48& 10.78&\textbf{\textcolor{gold}{28.06}}&\textbf{\textcolor{gold}{0.980}}&\textcolor{silver}{8.89}&\textbf{\textcolor{gold}{6.77}}\\
      \hline
      MonoSDF \cite{monosdf}         &HS& 8+8 &137.95& 6.35&25.52&0.965&13.16&9.70\\
            Ours (naive)             &TSDF&8+8& 7.80& 6.31&\textcolor{silver}{27.46}&\textcolor{silver}{0.978}&\textcolor{silver}{8.79}&\textcolor{silver}{6.98} \\
      Ours (full)              &TSDF& 8+8 & 7.80&5.68&\textbf{\textcolor{gold}{28.03}}&\textbf{\textcolor{gold}{0.980}}&\textbf{\textcolor{gold}{8.66}}&\textbf{\textcolor{gold}{6.82}}\\ 
      \hline
            MonoSDF \cite{monosdf}        &HS& 6+8& 183.93&  6.10 & 24.46 & 0.955& 15.87 & 11.07\\
            Ours (naive)             &TSDF& 6+8 &10.39& 6.12 & \textcolor{silver}{26.60}& \textcolor{silver}{0.973}&\textcolor{silver}{9.50}&\textcolor{silver}{7.18}\\
            Ours (full)             &TSDF& 6+8 &10.39& 5.26&\textbf{\textcolor{gold}{28.03}}&\textbf{\textcolor{gold}{0.980}}&\textbf{\textcolor{gold}{8.65}}&\textbf{\textcolor{gold}{6.85}}\\
            \hline
            MonoSDF \cite{monosdf}        &HS& 6+6 &183.93&  5.96&23.77&0.947&16.57&12.42 \\
            Ours (naive)             &TSDF& 6+6&10.39 & 5.97&\textcolor{silver}{26.49}&\textcolor{silver}{0.972}&\textcolor{silver}{9.23}&\textcolor{silver}{7.28} \\
            Ours (full)              &TSDF& 6+6 &10.39& 5.05&\textbf{\textcolor{gold}{28.00}}&\textbf{\textcolor{gold}{0.980}}&\textbf{\textcolor{gold}{8.50}}&\textbf{\textcolor{gold}{6.89}} \\ 
            \hline
            MonoSDF \cite{monosdf}        &HS& 3+4 &367.81& 5.39&20.90&0.897&19.61&16.75\\
        Ours (naive)             &TSDF& 3+4 & 20.78&5.48&\textcolor{silver}{24.55}&\textcolor{silver}{0.961}&\textcolor{silver}{11.91}&\textcolor{silver}{8.48}\\
            Ours (full)              &TSDF& 3+4&20.78 & 4.51&\textbf{\textcolor{gold}{27.80}}&\textbf{\textcolor{gold}{0.979}}&\textbf{\textcolor{gold}{8.55}}&\textbf{\textcolor{gold}{7.20}}\\ 
        \hline
        MonoSDF \cite{monosdf}        &HS& 3+2 &367.81 &4.91&18.37&0.836&127.63&17.30\\
            Ours (naive)             &TSDF& 3+2 &20.78& 5.05&\textcolor{silver}{22.99}&\textcolor{silver}{0.937}&\textcolor{silver}{16.30}&\textcolor{silver}{9.79}\\
            Ours (full)              &TSDF& 3+2 &20.78& 4.20&\textbf{\textcolor{gold}{27.36}}&\textbf{\textcolor{gold}{0.977}}&\textbf{\textcolor{gold}{7.96}}&\textbf{\textcolor{gold}{7.60}}\\
        \bottomrule
    \end{tabular}
    }
    \caption{Additional quantitative comparison on the Garage dataset. HS and EB stand for Hierarchical sampling~\cite{nerf,nerfinthewild,mipnerf,neus} and the Error bound sampling~\cite{volsdf}, respectively. The numbers of coarse and fine samples are combined with $+$. D [cm] indicates the depth error, measured by MAE. N [deg] is the normal error in degree.  
    % The average number of samples of Ours (full) method has been rounded off to integer for simplicity
    $\Delta t$ refers to the interval between coarse samples.}
\label{tab:suppl_garage}
\end{table*}
% Go below, and you will see some contents
% \end{appendices}

\end{document}